\pdfoutput=1
\documentclass[11pt]{article}

\usepackage{acl}

\usepackage{times}
\usepackage{latexsym}
\usepackage{multirow}

\usepackage{amsmath}
\usepackage{graphicx}
\usepackage{xcolor}

\usepackage[T1]{fontenc}
\usepackage[utf8]{inputenc}

\usepackage{microtype}

\usepackage{adjustbox}

\usepackage{inconsolata}

\usepackage{algorithm}
\usepackage{algpseudocode}
\usepackage{eqparbox}

\newlength{\myalgindent}
\algnewcommand\LeftComment[2]{%
  \hspace{#1\myalgindent}$\triangleright$ \text{#2}%
}

\title{A Data-Driven Guided Decoding Mechanism for Diagnostic Captioning}

\author{Panagiotis Kaliosis$^{1,2}$, John Pavlopoulos$^{1,2}$\thanks{Corresponding author.}, Foivos Charalampakos$^{1}$, \\ {\bf Georgios Moschovis$^{1,2}$}, {\bf Ion Androutsopoulos$^{1,2}$} \\ \\
         $^1$Department of Informatics, Athens University of Economics and Business, Greece \\ $^2$Archimedes/Athena RC, Greece \\
         \{pkaliosis, annis, phoebuschar, geomos, ion\}@aueb.gr}

\begin{document}
\maketitle
\begin{abstract}
Diagnostic Captioning (DC) automatically generates a diagnostic text from one or more medical images (e.g., X-rays, MRIs) of a patient. Treated as a draft, the generated text may assist clinicians, by providing an initial estimation of the patient's condition, speeding up and helping safeguard the diagnostic process. The accuracy of a diagnostic text, however, strongly depends on how well the key medical conditions depicted in the images are expressed. We propose a new \textit{data-driven} guided decoding method that incorporates medical information, in the form of existing tags capturing  key conditions of the image(s),  into the beam search of the diagnostic text generation process. We evaluate the proposed method on two medical datasets using four DC systems that range from generic image-to-text systems with CNN encoders and RNN decoders to pre-trained Large Language Models. The latter can also be used in few- and zero-shot learning scenarios. In most cases, the proposed mechanism improves performance with respect to all evaluation measures. We provide an open-source implementation of the proposed method at {\small\url{https://github.com/nlpaueb/dmmcs}}.

\end{abstract}

\section{Introduction}
Diagnostic Captioning (DC) systems receive one or more medical images of a patient, such as X-Rays or Magnetic Resonance Images (MRIs), which they analyse to draft a diagnostic report 
% describing the medical conditions of that patient 
\cite{dc-survey-ting}. 
%is a task within the broader domain of Image Captioning, which has 
%experienced increasing research attention in the past few years \cite{dc-survey-ting}. 
%DC systems , and aim to generate a draft diagnostic report that outlines the key medical findings regarding the patient's condition. 
Such systems can function as supportive tools for doctors and clinical staff, assisting them in their daily workload. Possible benefits, as summarized by \citet{Pavlopoulos2021DiagnosticCA}, include  (\romannumeral 1) increased overall throughput of medical departments, since improving a partially correct draft report may be faster than writing it from scratch, (\romannumeral 2) reduced diagnostic errors, by providing suggestions for the clinical findings of the input images, which might otherwise be missed, and (\romannumeral 3) decreased cost of medical imaging examinations.
%\cite{deepl-mi-general-overview}.
Despite the rapid advancements in deep learning methods,  draft diagnostic reports generated by DC systems still exhibit shortcomings, such as hallucinations or lack of accurate descriptions of
% the conditions depicted in the images 
medical findings \cite{deep-image-captioning-survey}. 
% Although the former can be partly addressed with larger language models,
%% , even though they also hallucinate sometimes
The medical accuracy of a generated diagnostic text strongly depends on whether key medical conditions depicted in the images are considered during text generation \cite{Huang2019MultiAttentionAI, Wang2020UnifyingRS}. Such key conditions can be captured by tags, reflecting medical concepts to be mentioned in the generated text. Tags of this kind can be obtained by medical image taggers \cite{Rajpurkar2017CheXNetRP, lu2020semi} and are also present (as gold tags) in diagnostic captioning datasets.
% terms can be retrieved from the patient's medical history or generated by a medical image classifier and are widely referred to as medical ``concepts'' or ``tags'' \cite{pavlopoulos-etal-2019-survey2}. 
Assigning tags to an image is to some extent similar to content selection, i.e., deciding
% which tokens to generate at a conceptual or logical level 
which concepts to express, which was the first stage in  symbolic-based text generation systems \cite{reiter_dale_2000}. More background information on DC is provided in Appendix \ref{sec:dc-background}, and in the DC survey by \citet{Pavlopoulos2021DiagnosticCA}.

In this work, we propose \emph{Distance from Median Maximum Concept Similarity} (DMMCS), a novel data-driven guided decoding method that aims to integrate information from medical image tags into the diagnostic text generation process. This is achieved by imposing a new penalty at each decoding step. The penalty is designed to prioritize the generation of words that are semantically similar to the medical tags of the input images, also taking into account how often each tag is explicitly or implicitly expressed in gold captions. 
% In this way, we manage to integrate information from key biomedical terms regarding the patients' condition into the generated diagnostic reports. Thus, the semantic affinity, i.e., the alignment or resemblance in meaning, between the generated and the ground-truth report can be enhanced.
DMMCS involves calculating a series of statistical distributions that model the relationship between each tag and the tokens of the diagnostic captions it is associated with in the training data. 
% To the best of our knowledge, this 
It is the first guided decoding method specifically developed for DC, as well as the first \emph{data-driven} method that uses image tags to guide the generation of image captions.

DMMCS is applicable to any encoder-decoder DC system, as it can be integrated in the decoding process. For experimental purposes, we train four DC systems, ranging from a generic CNN-RNN image-to-text method \cite{Vinyals2014ShowAT}, to Transformer-based architectures \cite{Vaswani2017AttentionIA}, and state-of-the-art prompt-based systems \cite{instructblip, alayrac2022flamingo}. Furthermore, we investigate the impact of DMMCS on few-shot captioning scenarios. We evaluate the performance of all models on two medical datasets,  ImageCLEFmedical 2023 \cite{ImageCLEFmedicalCaptionOverview2023} and MIMIC-CXR \cite{mimic-cxr}. We use two evaluation measures, BLEU and BLEURT, comparing the results with and without DMMCS, demonstrating the effectiveness of the proposed algorithm. As one would expect, the performance boost provided by DMMCS is larger when using gold tags, but we show that DMMCS is also beneficial with noisy tags predicted by medical image classifiers. 

Our main contributions are: (\romannumeral 1) We introduce DMMCS, a data-driven guided decoding method for DC, which leverages medical tags of the input images to improve the generated captions.
(\romannumeral 2) We incorporate DMMCS in four DC systems covering a range of learning scenarios, including fine-tuning and few-shot learning. (\romannumeral 3) We demonstrate that DMMCS significantly enhances performance across all four models in most cases, even when using noisy predicted tags.

\section{Related Work}

Substantial research has been dedicated to Controllable Text Generation and guided decoding strategies \cite{prabhumoye-etal-2020-exploring2}. Standard decoding methods, such as greedy or standard beam search, provide minimal control over the model's output \cite{pmlr-v202-zhou23g}. Therefore, decoding techniques that partially guide the model's choices to adhere to task-specific requirements have been proposed. Most guided decoding methods can be categorized based on three control conditions: semantic, structural, and lexical \cite{Zhang2022ASO}. 

Semantic constraints guide the model to generate text conforming to specific attributes such as tense or sentiment \cite{yang-style-transfer, gu-etal-2022-distributional2}. For instance, \citet{ghazvininejad-etal-2017-hafez2} introduced \textit{Hafez}. It was initially designed for generating poetry-styled texts, but is adaptable to meet any specified content-based decoding constraint. This is achieved by integrating a constraint-specific score in the next word selection process at each decoding step. The score is calculated by a feature function. Such functions can, for example, encourage or discourage specific word choices, prevent repetitions, or guide the model to prefer longer words. The constraints may also be based on supervised learning \cite{holtzman-etal-2018-learning2} or reinforcement learning \cite{li2017learning}, rather than just heuristics \cite{baheti-etal-2018-generating2}. 

Structural constraints direct the model's output to adhere to a specific syntax structure \cite{yang-etal-2022-diversifying2, chen-etal-2019-controllable2, kumar-etal-2020-syntax2}. Finally, lexical-based constraints guide the model's choices to incorporate a set of specified keywords. \citet{he-2021-parallel2} introduced CBART (constrained BART), which shifts a part of the generation process from the decoder to the encoder. The encoder guides the generation towards some specified must-include tokens. This is achieved by adding a dense layer over BART's encoder \cite{lewis-etal-2020-bart2} that generates a sequence of labels. The latter guides the decoder on which actions should be taken. The model undergoes a refinement process, regenerating multiple outputs until all constraints are fulfilled.
% Other extensions of guided decoding include Discriminative Adversarial Search \cite{scialom2020discriminative2}, learning a committee of discriminators specialized in different principles of communication to guide generation with RNNs \cite{holtzman-etal-2018-learning2}, or beam search as a regularized decoding framework \cite{meister-etal-2020-beam2}.

A recent related method for semantic guided decoding, which comprised the starting point of our research, is \textit{Contrastive Search} \cite{su2022a}. 
It attempts to tackle the problem of text degeneration, where models produce unnatural and repetitive text. It imposes a degeneration penalty at each decoding step to guide the model's output towards more natural and less repetitive text sequences. The penalty for a candidate token $u$ at decoding step $t$ is:
\begin{align*}
\label{eq1}
\tag{1}
D_u = \max\limits_{1 \le j \le t-1} {\textit{sim}\left(h(u), h(x_j)\right)},
\end{align*}

\noindent where $x_j$ are the preceding tokens of the incomplete generated sequence $x$, $h\left(\cdot\right)$ calculates word embeddings, and $\textit{sim}(\cdot, \cdot)$ denotes cosine similarity.
% function, i.e., $\textit{sim}(x, y) =   \frac {x \cdot y} {\left\| x\right\|\left\| y\right\|} $. 
The degeneration penalty $D_u$ is the maximum cosine similarity between the word embeddings of the candidate token $u$ and the preceding sequence $x_{<t}$.
% (denoted as $h(u)$ and $h(x_t)$ respectively). 
It is subtracted from the score that the decoder would otherwise assign to $u$, thus guiding the decoder to generate less repetitive text.

\section{The Proposed DMMCS Method}
\label{sec:method}

Our proposed method introduces a tag-guided decoding strategy for DC systems. It aims to guide the model to select words that appropriately express the tags (medical concepts) associated with the input image.\footnote{In our experiments, the input is always a single image (and its tags), but DMMCS also applies to multi-image inputs.} 
% This is achieved by imposing a custom penalty at each decoding step. 
% The tags outline the key aspects of the patient's medical condition. Hence, integrating information from them into the diagnostic caption can lead to enhanced accuracy of the draft report. 
For instance, if a radiology image is associated with the tag ``Atelectasis'', but the generated caption makes no implicit or explicit reference to the aforementioned medical condition, then it is probably inaccurate. 

\begin{figure}[th!]
    \centering
    \includegraphics[width=\linewidth, height=0.5\linewidth]{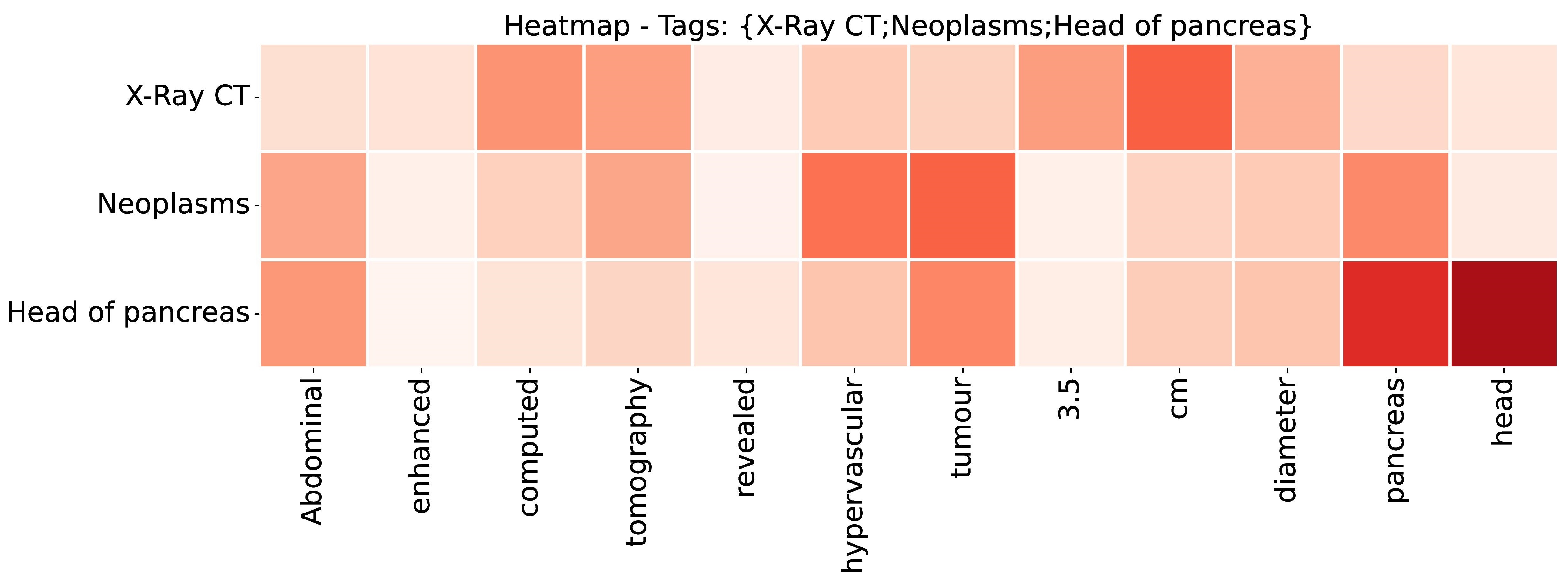}
    \vspace{-3mm}
\caption{Heatmap visualizing the cosine embedding similarities between the tokens of a ground truth caption (x-axis) and its associated biomedical tags (y-axis).}
\label{fig:3.1} 
\end{figure}

% As a first step, we conducted an exploratory analysis on the training set of the ImageCLEFmedical 2023 dataset \cite{ImageCLEFmedicalCaptionOverview2023}, in order to explore the relationship between the biomedical tags and their associated captions. 
As a first exploration, we calculated the FastText word embeddings \cite{fasttext} of all the tags and all the tokens of the gold captions in the training set of the ImageCLEF 2023 dataset \cite{ImageCLEFmedicalCaptionOverview2023}. Tags consisting of multiple tokens were represented by the centroid of the tokens' embeddings. We then investigated the relationship between each tag and the gold captions it was associated with (the gold captions of training images tagged with the particular tag). Figure \ref{fig:3.1} presents a heatmap that visualizes the relationship between the tokens of a caption $s$ (\textit{x}-axis) and its corresponding set of tags $T$ (\textit{y}-axis). Each heatmap cell represents the cosine similarity between the (centroid) embedding of the corresponding tag $t$ and the respective caption token $s_j$. Darker colors correspond to larger similarity values.
Hence, for instance, the lower right cells show that tokens $s_{11}$ and $s_{12}$ (``pancreas head'') of the caption have a very high cosine similarity with tag $t_3$ (``Head of pancreas''). Indeed, $t_3$ is almost explicitly expressed in $s$ (almost the same words), while the other two tags are expressed more implicitly (with different words, e.g., ``Neoplasms'' vs.\ ``tumour'', ``X-Ray CT'' vs.\ ``computed tomography''). We define the similarity between a tag $t$ and a caption $s$ as the \textit{maximum cosine similarity (MCS)} between the (centroid) word embedding of $t$ and the word embedding of each token in $s$, i.e.,
% . Mathematically, MCS is calculated as:
\begin{align*}
\label{eq2}
\tag{2}
    \textit{MCS}(t, s) = \max\limits_{1 \le j \le |s|} {\textit{sim}(h(t), h(s_j))}.
\end{align*}
% \noindent where $\textit{sim}(\cdot, \cdot)$, $h\left(\cdot\right)$ 
% % denote the cosine similarity and the embeddings generation function respectively, as defined 
% are as in Eq.~\ref{eq1}. 
A high $\textit{MCS}(t, s)$ between a tag $t$ and a caption $s$ indicates a significant presence of the tag's meaning in the caption.
Next, we investigated the relationship between each tag $t$ and the set $S$ of all the gold captions tag $t$ is associated with in the training data (gold captions of images tagged with $t$). For each tag $t$ and its associated captions $S$, we compute the  distribution $R(t, S)$ of its corresponding MCS scores, as the set:
\begin{align*}
\label{eq3}
\tag{3}
    {R(t, S) = \text{\{}\textit{MCS}(t, s) | s \in S\text{\}}}.
\end{align*}
% \noindent where $\textit{MCS}(t, s)$ is defined in Eq.~\ref{eq2}.

\begin{figure}[t]
    \centering
    \includegraphics[width=.30\textwidth]{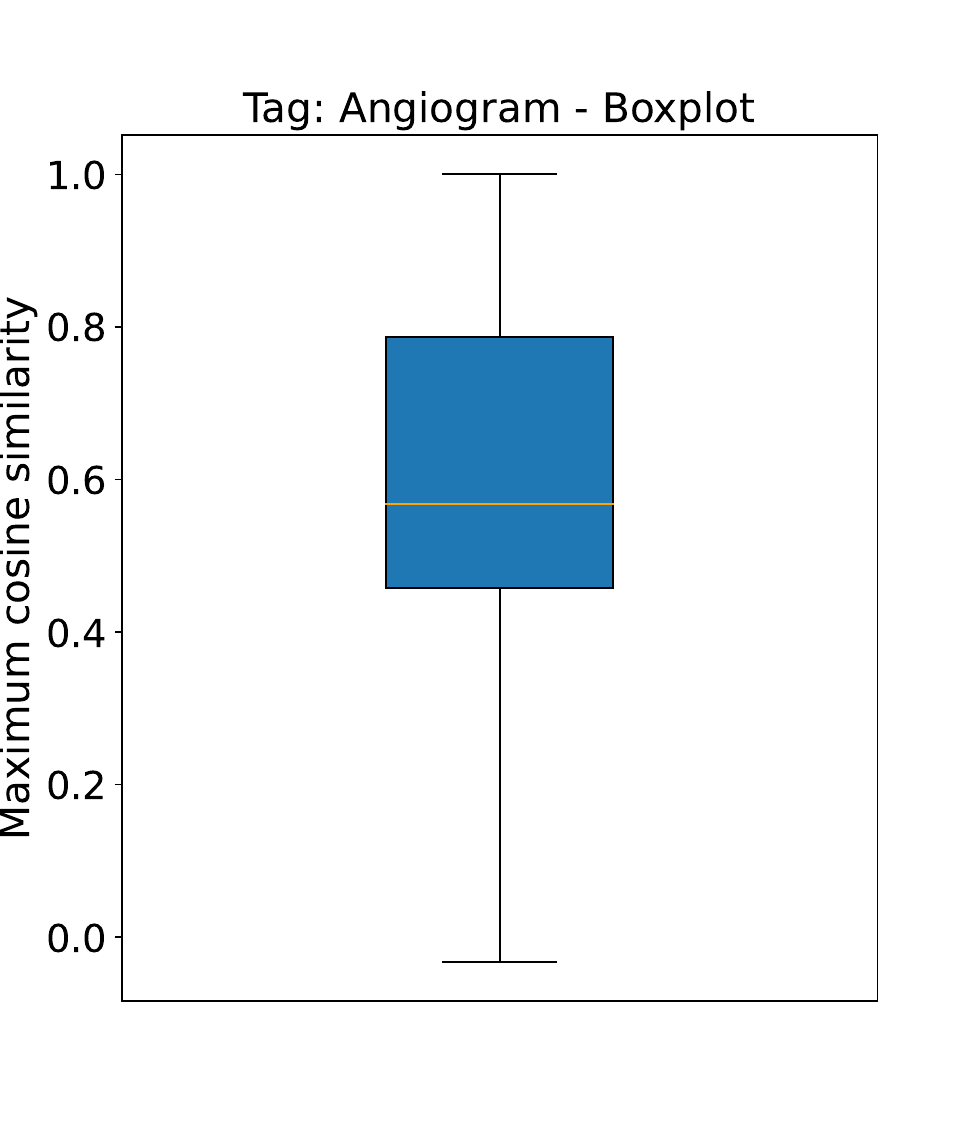}
    \vspace{-5mm}
\caption{Distribution $R(t, s)$ for the tag ``Angiogram''. Interquartile range (IQR) shown as blue box. The coral line is the median, denoted $\textit{MMCS}(t, S)$.}
\vspace{-5mm}
\label{fig:3.2} 
\end{figure}

\noindent Figure \ref{fig:3.2} illustrates the distribution $R(t, S)$ of the tag ``Angiogram'', using the gold captions of the images associated with this particular tag in the ground truth of the ImageCLEF 2023 training set. The blue box represents the interquartile range (IQR) of the distribution, while the coral line denotes the median value, denoted $\textit{MMCS}(t)$: %based on the training data.
\begin{align*}
\label{eq4}
\tag{4}
    \textit{MMCS}(t) = \textit{median}(R(t, S)).
\end{align*}

\begin{figure}[h!]
    \centering
    \includegraphics[width=8.5cm, height=6.5cm, trim={2.3cm 0cm 0cm 3.5cm}, clip]{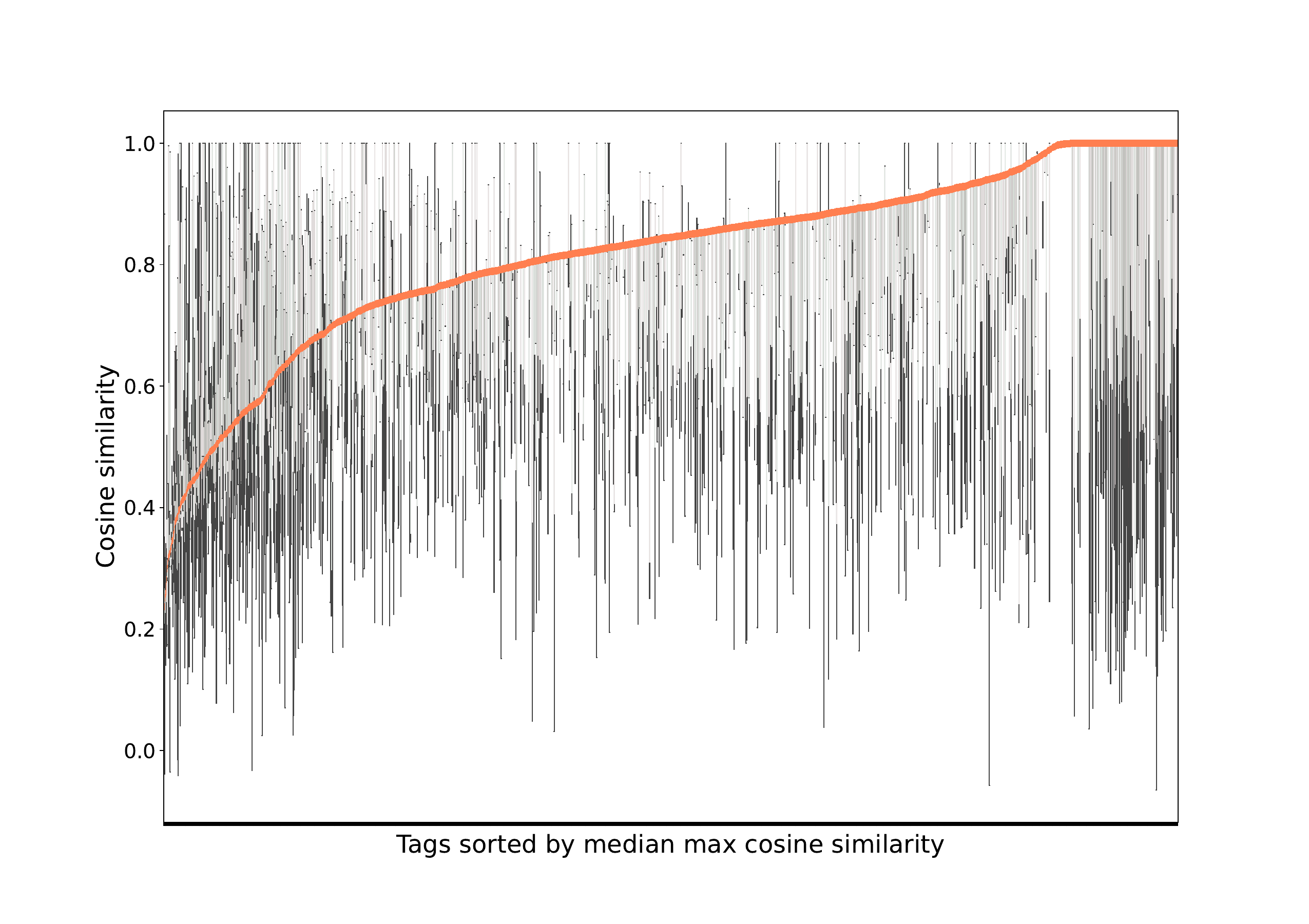}
    \vspace*{-8mm}
\caption{$\textit{MMCS}(t, S)$ (coral line) and IQR (vertical lines) per tag $t$, sorted by ascending $\textit{MMCS}(t, S)$.}
%figs/mmc-bnw-4.pdf
\vspace{-2mm}
\label{fig:3.3} 
\end{figure}

We repeated the calculation of the distribution $R(t, S)$ for every tag $t$ and the set $S$ of gold captions associated with $t$ in the training set of ImageCLEF 2023. Figure \ref{fig:3.3} shows the interquartile range (IQR, black vertical lines) and $\textit{MMCS}(t, S)$ (median, coral line) of the distribution $R(t, S)$ for each tag $t$. In other words, Figure~\ref{fig:3.3} contains many box-plots, like the one of Figure~\ref{fig:3.2}, side by side. Intuitively, Figure~\ref{fig:3.3} shows how strongly each tag $t$ is expressed in the gold captions associated with it. The side-by-side box-plots of Figure~\ref{fig:3.3} are sorted by ascending $\textit{MMCS}(t, S)$ (coral), in order to highlight the observation that the tags of the ImageCLEFmedical 2023 dataset are not expressed equally strongly in the ground truth captions.
% (see also \S\ref{sec:experiments}). 
For instance, we observe a variation in $\textit{MMCS}(t, S)$ values (coral line), ranging from as low as $0.3$ for tags on the left end, to values close to $1$ for tags on the right end. The former are overall expressed more implicitly in the diagnostic captions they are associated with, while the latter are explicitly mentioned. Some tags conveying information that may be trivial to a clinician (e.g., that the image is an X-Ray) may actually not be expressed at all (not even implicitly).

Considering these findings, we propose a new decoding penalty that aims to improve the generated diagnostic captions by integrating information provided by the image's medical tags. The tags are in practice predicted by a medical image tagger, but we experiment with both predicted and gold (oracle) tags. The penalty encourages the decoder to select words that express more or less explicitly (or not at all) the tags of the image. The target level of explicitness of each tag is determined by its $\textit{MMCS}(t, S)$ score, which is computed on the training captions. Tags with larger $\textit{MMCS}(t, S)$ should be explicitly mentioned, while tags with lower $\textit{MMCS}(t, S)$ should be expressed less explicitly (or not at all).

% Regarding the definition of the penalty, 
During inference, if an image is associated with a single tag $t$, we calculate the $\textit{MCS}(t, s)$ (Eq.~\ref{eq2}) between the tag $t$ and the tokens of each candidate (possibly still incomplete) caption $s$ being considered by the beam search decoder. The penalty is the squared difference between the computed $\textit{MCS}(t, s)$, which shows how strongly the tag is expressed in the candidate caption, and the tag's $\textit{MMCS}(t, S)$, which shows how strongly the tag is expressed (median value) in the ground-truth training captions it is associated with. If an image is associated with multiple tags, then a separate  penalty is calculated for each tag, as above, and the total penalty is the sum of the penalties divided by the number of associated tags. Formally, given a (possibly incomplete) caption $s$ and a set of image tags $T$ to be expressed, the penalty is calculated as:
\begin{align*}
\small
    \textit{DMMCS}_{\textit{p}}(T,s) & = \\
    \frac{1}{|T|} \cdot & {\sum_{t \in T}(\textit{MCS}(t,s) -  \textit{MMCS}(t))^ 2}.
\label{eq5}
\tag{5}
\end{align*}
%
% In essence, the penalty favors a candidate caption $s$ whose $\textit{MCS}(t, s)$ (computed across the tokens of $s$) is closer to $\textit{MMCS}(t, S)$, for all the tags $t$ associated with the image. The former reflects the maximum semantic representation of the tag in the words of the candidate caption, while the latter denotes the average semantic representation of the tag in the captions it is associated with in the training set. 
Intuitively, the goal is to generate a caption that expresses each tag of the image as strongly 
% (close to the median strength) 
as the training captions associated with the tag.

At each decoding step, each candidate (possibly incomplete) caption $s$ considered by the beam search decoder is scored as follows:
\begin{align*}
    \textit{DMMCS}(s) = & \ \alpha \cdot \textit{DMMCS}_{\textit{p}}(T,s) \\ 
    & + (1-\alpha) \cdot (1-\textit{D}_{\textit{score}}),
\label{eq6}
\tag{6}
\end{align*}

\noindent where $T$ is the set of tags the input image  is associated with. $\textit{D}_{\textit{score}}$ and $\alpha$ are explained below.

\smallskip\noindent \textbf{$\textit{D}_{\textit{score}}$:} This is the score the decoder assigns to each candidate caption, i.e., the sum of the (log) probabilities of the decoder (Eq.~\ref{eq7}). The decoder can be any type of generative model conditioned on the input image, e.g., a Recurrent Neural Network (RNN) or a Transformer-based architecture \cite{Vaswani2017AttentionIA}. We conducted experiments with both types of architectures in order to test the effectiveness of the DMMCS-based decoding. 
$\textit{D}_{\textit{score}}$ is also normalized in $[0,1]$, using min-max scaling, in order to align with the score range of $\textit{DMMCS}_{\textit{p}}$. As the goal is to minimize the overall score, Eq.~\ref{eq6} uses $1  - \textit{D}_{\textit{score}}$.
\begin{align*}
\label{eq7}
\tag{7}
    \textit{D}_{\textit{score}} = -\sum_{t=1}^{t} \log P(s_t | s_{<t})
\end{align*}

\noindent \textbf{$\alpha$:} This hyper-parameter controls the effect of the two terms in the overall $\textit{DMMCS}(s)$ score. The larger the value of $\alpha$, the more significant the influence of the penalty $\textit{DMMCS}_{\textit{p}}$ on the overall score, and vice-versa. When $\alpha=0$, standard beam search is applied.
% , while when $\alpha=1$, we solely trust the DMMCS mechanism. 
The value of $\alpha$ is tuned by experimenting with several values on the validation set. 
% It is also used in order to scale the value of the two addends, in order to ensure that both will substantially contribute to the overall score. 

In summary, at each decoding step the $\textit{DMMCS}(s)$ score (Eq.~\ref{eq6}) is calculated for each candidate sequence $s$ of the beam search. The score combines $\textit{DMMCS}_{p}$ (Eq.~\ref{eq5}) and $\textit{D}_{score}$ (Eq.~\ref{eq7}). The former measures how well the input image's tags are expressed in the candidate (possibly incomplete) caption, while the latter denotes the score assigned by the decoder to the candidate caption. The $n$ sequences with the best $\textit{DMMCS}(s)$ scores are selected and expanded at the next decoding step. Algorithm \ref{alg:cap} presents our proposed guided decoding method in pseudo-code form.

\newcommand\Algphase[1]{%
  \vspace*{-.1\baselineskip}\Statex\hspace*{0pt}\rule{\linewidth}{0.4pt}%  
  \Statex\hspace*{\algorithmicindent}\textbf{#1}% 
  \vspace*{-.1\baselineskip}\Statex\hspace*{0pt}\rule{\linewidth}{0.4pt}%
}

\begin{algorithm}
\caption{DMMCS algorithm - Pseudocode}\label{alg:cap}
\begin{algorithmic}
\State $T, S, I$ \Comment{the dataset tags, captions, images}
\State $i \in I$ \Comment{an image in question}
\State $\text{tags}_i^p, \text{tags}_i^g \subseteq T$ \Comment{predicted, gold tags of $i$}

\Algphase{Phase 1 - Training Statistics}
%\State $\text{beamScores}$ \Comment{score per beam}
\For{$t \in T$}
\State $S'$ $\leftarrow$ $\{s_i \in S: t \in \text{tags}_i^g\}$ \Comment{captions per $t$}
% \State $\text{MCS}_t \gets \text{calculateMCS}(t, S')$
% \Comment Alg. ~\ref{euclid}

\vspace{0.2em}

\State $\text{MMCS}(t) \gets \text{median}\{\text{MCS}(t, s) | s \in S'\}$
\State \Comment Eqs.\ \ref{eq2}--\ref{eq4}
% \textit{median}(\text{MCS}_t)$
\EndFor
% \Algphase{Phase 2 - Inference}
% \For{\textbf{each} decoding step of the beam search}
% \For{\textbf{each} beam search sequence $s$}
% \State \text{get} $D_{score}$ \Comment{decoder's score of $s$}
% %\State $\text{sum} \gets 0$
% \vspace{0.2em}
% % \For{$t \in \text{tags}_i^p$}
% % \State $\text{MCS}_{t'} \gets \text{calculateMCS}(t', s)$ 
% % \State $\text{sum} \gets \text{sum} + (\text{MCS}_{t'} - \text{MMCS}_{t'})^2$
% % \EndFor
%     \State \LeftComment{2}{Calculate DMMCS penalty (Eq.~\ref{eq5})}
%     \State $T'$ \gets \text{tags}_i^p \Comment{predicted tags of $i$}
%     \State \text{DMMCS}_{\textit{p}} \gets
%     \frac{1}{|T'|}
%     \sum_{t \in T'}(\text{MCS}(t,s) -\text{MMCS}(t))^2
%     \State  
%     \vspace{0.2em}
%     % \State $\text{DMMCS}_p \gets \text{sum} / |\text{tags}_i^p|$
%     \vspace{0.4em}
%     \State \LeftComment{2}{Calculate DMMCS score (Eq. ~\ref{eq6})}
%     \State $\begin{aligned}
%     \text{DMMCS}_s \gets & \alpha \cdot \text{DMMCS}_p \\
%     & + (1 - \alpha) \cdot (1 - D_{score})
%     \end{aligned}$

%     \vspace{0.4em}
%     \State \LeftComment{2}{Set updated beam scores}
%     \State $\text{beamScores[$s$]} \gets \text{DMMCS}_s$
% \EndFor
% \EndFor

\Algphase{Phase 2 - Inference}
\For{\textbf{each} decoding step of the beam search}
    \For{\textbf{each} beam search sequence $s$}
        \State \text{get} $D_{score}$ \Comment{decoder's score of $s$}
        
        \vspace{0.6em}
        
        \State \LeftComment{2}{Calculate DMMCS penalty (Eq.~\ref{eq5})}
        
        \State $T' \gets \text{tags}_i^p$ \Comment{predicted tags of $i$}

        \vspace{0.2em}
        \State $\begin{aligned}
             \text{DMMCS}_{\textit{p}} \gets \frac{1}{|T'|} \sum_{t \in T'}(\text{MCS}(t,s) - \\
            \text{MMCS}(t))^2
        \end{aligned}$

        \vspace{0.6em}
        \State \LeftComment{2}{Calculate DMMCS score (Eq.~\ref{eq6})}
        \State $\begin{aligned}
            \text{DMMCS}_s \gets & \alpha \cdot \text{DMMCS}_p \\
            & + (1 - \alpha) \cdot (1 - D_{score})
        \end{aligned}$
        
        \vspace{0.6em}
        \State \LeftComment{2}{Set updated beam scores}
        \State $\text{beamScores[$s$]} \gets \text{DMMCS}_s$
    \EndFor
\EndFor

\end{algorithmic}
\end{algorithm}

\section{Experiments}
\label{sec:experiments}

% In this section, we discuss the experiments conducted to assess and validate the effectiveness of the DMMCS penalty for DC. 
For experimental purposes, we employed DMMCS decoding on four DC models, spanning diverse architectures and learning techniques, including long-established generic image captioning models and more advanced state-of-the-art ones. Moreover, we evaluated DMMCS using two  datasets: the dataset of ImageCLEFmedical 2023 \cite{ImageCLEFmedicalCaptionOverview2023}, and a subset of the MIMIC-CXR dataset \cite{mimic-cxr}. The performance of each model is evaluated using two measures, briefly discussed in \S\ref{sec:4.3}. We demonstrate that DMMCS leads to substantial performance improvements in most cases.

\subsection{Datasets} \label{sec:datasets}

%Here we briefly present the two datasets used for the evaluation of the proposed algorithm.
%\\

\noindent\textbf{ImageCLEFmedical 2023 Dataset:} This dataset \cite{ImageCLEFmedicalCaptionOverview2023} is a subset of the Radiology Objects in Context (ROCO) dataset \cite{ROCO}. It consists of $71{,}355$ images, accompanied by their gold tags and gold diagnostic captions. The maximum number of words in a single caption is $315$, the minimum is $1$, while the average is $16.04$.
% , as well as a brief diagnostic caption. 
The dataset contains $2{,}125$ distinct tags, and the images are from multiple medical modalities, such as X-Ray, CT, MRI. No ground truth captions for the test set are available. We split the provided data into three subsets, following a 75\%-10\%-15\% split, holding out a test subset for evaluation purposes. Thus, we employed $53,516$ images as our training data, $7,135$ images as our validation set, while the remaining $10,704$ images comprised our held-out test set.

\smallskip\noindent\textbf{MIMIC-CXR Dataset:} MIMIC-CXR \cite{mimic-cxr} originally consists of $377{,}110$ deidentified chest X-ray examinations. Each of them is accompanied by a set of gold medical tags and a gold 
% brief 
diagnostic caption. The maximum number of words in a single caption is $1,014$, the minimum is $2$, while the average is $137.8$. There are $14$ distinct biomedical tags in total, generated by the CheXpert labeler system \cite{chexpert}. We only considered examinations containing a single radiology image per patient, because the DC systems we used could only receive a single image as input. Thus, we used $32{,}637$ images, along with their gold tags and captions. We randomly split them into training, validation and test sets, containing $25{,}301$, $2{,}752$ and $4{,}584$ samples, respectively.

\subsection{DC Models}

% The following four DC Models were used for the evaluation of the proposed algorithm.

\noindent\textbf{Show \& Tell:} This is a basic generic image captioning model 
% is known as Show \& Tell architecture 
\cite{Vinyals2014ShowAT}, 
% or CNN-RNN, as it 
consisting of a 
% Convolutional Neural Network (CNN) 
CNN image encoder followed by an RNN text decoder (no attention used).\footnote{Show, Attend \& Tell \cite{showattendtell} with attention 
%has been reported to perform worse in DC \cite{Zachariadis2022}.} 
performed worse in DC in preliminary experiments.} 
The CNN extracts image features that are then fed to the RNN decoder, which generates the diagnostic caption word by word, based on the image features.

\smallskip\noindent\textbf{ViT-GPT2:} This is also based on the encoder-decoder approach, but now both components are based on Transformers \cite{Vaswani2017AttentionIA}. Vision Transformer (ViT) \cite{vit} is employed as the image encoder, while GPT2 \cite{gpt2} is responsible for caption generation. Both models were loaded from a HuggingFace checkpoint\footnote{\tiny\url{https://huggingface.co/nlpconnect/vit-gpt2-image-captioning}} for a joint ViT-GPT2 encoder-decoder pipeline, and were further pre-trained on generic image-caption pairs. We then fine-tuned the model on the two employed DC datasets.

\smallskip\noindent\textbf{InstructBLIP:} This is vision-language instruction-tuned model \cite{instructblip}. Such models are designed to swiftly adapt to new tasks based on specific instructions.
% provided during the training process. 
Hence, their performance strongly depends on the provided instructions (presented in Appendix~\ref{sec:instructions}).

\smallskip\noindent\textbf{Flamingo:} This is a few-shot (in-context learning) generic image captioning system \cite{alayrac2022flamingo}. Flamingo can generate a diagnostic caption based on a few demonstrative examples of image-caption pairs provided as a multi-modal prompt. Checkpoints of the original Flamingo architecture are not publicly available, hence we employed OpenFlamingo, an open-source implementation that obtains around $80$-$89\%$ of the original 
% DeepMind's 
Flamingo's performance \cite{openflamingo}.

\subsection{Evaluation Measures}
\label{sec:4.3}

%We provide a brief introduction of the two evaluation measures we use in order to assess the proposed algorithm: 
\textbf{BLEU, BLEURT:} 
We use BLEU \cite{papineni-etal-2002-bleu2} and the more recent (BERT-based) BLEURT \cite{sellam-etal-2020-bleurt2} as main evaluation measures. Both are frequently used in text generation, including generic image-to-text and DC.\footnote{We also report ROUGE scores 
% a recall-based measure, 
in Appendix \ref{sec:classifier}.}
% BLEU is based on $n$-grams and evaluates the lexical overlap between a machine-generated text and a ground-truth text, prioritizing precision (checking if generated $n$-grams are included in the ground-truth). Precision is particularly important in DC due to the critical nature of medical information. It refers to the accuracy of the generated captions in conveying the intended meaning, particularly in terms of using the correct medical terminology and describing the medical condition accurately. Inaccuracies or errors in the generated captions could lead to misinterpretations, potentially affecting the medical decision-making process. 
% \noindent On the other hand, BLEURT \cite{sellam-etal-2020-bleurt2} is a state-of-the-art evaluation metric designed to approximate human-level evaluation by leveraging BERT's natural language understanding capabilities. It undergoes a two-stage pre-training process designed to equip it with foundational knowledge across a range of text evaluation tasks. In the first stage, it is trained on a series of typical text evaluation tasks, including BLEU. Consequently, it is fine-tuned on a smaller human-annotated dataset consisting of generated text and corresponding human judgments. In this way, BLEURT learns to assign scores that indicate the degree of similarity between the generated and reference texts, aligning closely with human assessments of text quality.

% \subsection{Clinical Correctness}
\label{sec:4.4}
\smallskip\noindent\textbf{Clinical Accuracy:}
Previous work has shown measures like BLEU and BLEURT  may not adequately capture clinical correctness in DC 
% between the generated and reference sequences 
\cite{Pavlopoulos2021DiagnosticCA}. Hence, we also measure  \emph{clinical accuracy} (CA) by comparing medical concepts extracted from the generated diagnostic captions (as silver labels by a multi-label text classifier) to those extracted from the corresponding gold captions.
% We use CheXpert \cite{Irvin2019CheXpertAL} to extract medical terms, resulting into a binary vector for each text. We then proceed to calculate the accuracy score achieved on each category. The aggregate score across all texts is determined by the macro-averaged accuracy scores of individual categories.
To compute CA, we follow previous work \cite{pmlr-v106-liu19a} that employs CheXBERT \cite{Irvin2019CheXpertAL} to determine whether or not each one of the 14 thoracic ailments (treated as classes) are mentioned in a given diagnostic caption. Given a caption (generated or gold), CheXBERT produces “present”, “negative”, “unsure” or “blank”, for each one of the 14 ailments. We treat “blank” as “negative”. When the “unsure” label is predicted, we change the label to either “present” or “negative” with equal probability, as in the work of \citet{pmlr-v106-liu19a}. 
% We first calculate the accuracy score achieved on each category and the aggregate score across all texts is determined by the macro-averaged accuracy scores of individual categories.
CA is defined as:
\[ \textit{CA} = \frac{1}{m} \sum_{j=1}^{m} \frac{1}{n} \sum_{i=1}^{n} 1\{y_{R}[i, j] = y_{P}[i, j]\}\text{,}\]

\smallskip\noindent where $n$ is the number of classes, $m$ is the number of caption pairs (gold-generated) being compared, and $y_{R}, y_{P}$ denote predictions from reference (gold) and predicted (generated) captions, respectively. 
% We measure CA only on MIMIC-CXR, which shares the classes (medical concepts) with CheXBERT, unlike ImageCLEFmedical 2023 where CheXBERT is not applicable.
% ensuring a direct comparison between the generated and reference captions.

\subsection{Baselines}
\label{sec:baselines}

% We compare DMMCS against two baselines.
% decoding approaches.
% commonly used in evaluations of guided decoding tasks. 

\smallskip\noindent\textbf{Beam search decoding:} The first baseline we compare DMMCS against is standard beam search. 

\smallskip\noindent\textbf{Constrained beam search:} The second baseline is a constrained beam search decoding method \cite{anderson-etal-2017-guided}, where decoding is guided by a set of manually defined constraints, aiming to enforce specific lexical requirements in the generated captions. We experimented with two versions of constrained beam search, namely strict and disjunctive. The former (denoted $\forall$) ensures that all specified keywords are present in the generated captions. The latter (denoted $\exists$) only enforces the inclusion of at least one of the given tags in the generated captions. For both versions, we used the HuggingFace implementation of constrained beam search,\footnote{\tiny\url{https://huggingface.co/blog/constrained-beam-search}} which is based on a plethora of guided decoding methods \cite{anderson-etal-2017-guided, post-vilar-2018-fast, hu-etal-2019-improved, li-etal-guided}. 
% By comparing our method with the aforementioned baselines, we aim to demonstrate its, in most cases, superior performance when generating accurate and clinically relevant diagnostic captions.

\subsection{Experimental Results}
\label{sec:results}

% In this section, we present a quantitative and qualitative analysis on the results of our experiments with the proposed decoding algorithm. Moreover, we assess each model's performance both with and without the integration of the DMMCS-based decoding. 
%We repeat each experiment for three random non-intersecting subsets of the test set (each containing $1000$ image samples) and report the average score for each evaluation measure. 
We repeat each experiment for three random non-intersecting subsets of the test set, with each subset containing $1,000$ images. 
% By evaluating the model's performance across multiple random splits, we ensure robustness and reduce the impact of any specific composition of the test set. 
We report the average (over the the three test subsets) score for each evaluation measure and the 
% variance, 
standard deviation, 
offering insights into the stability of each model's performance across different test subsets.
To obtain the medical tags of each image, we use a medical image tagger trained on the training subset of each dataset. Specifically, we employ the top-performing encoder of the ImageCLEFmedical 2023 campaign \cite{ImageCLEF2023, ImageCLEFmedicalCaptionOverview2023}, namely a DenseNet-121 instance \cite{Huang2016DenselyCC}, initially pre-trained on ImageNet \cite{deng2009imagenet}. We fine-tuned it (separately per DC dataset) on the training images and corresponding gold tags of the two DC datasets we use (Section~\ref{sec:datasets}).
% biomedical domain using the respective dataset's image-tags pairs. 
We also report scores using the gold (oracle) tags of the test images (Appendix~\ref{sec:classifier}).

\begin{figure}[t]
    \centering
    %\begin{tabular}[b]{c}
    \includegraphics[width=\linewidth, trim={0.5cm 3cm 0.6cm 2.75cm}, clip]{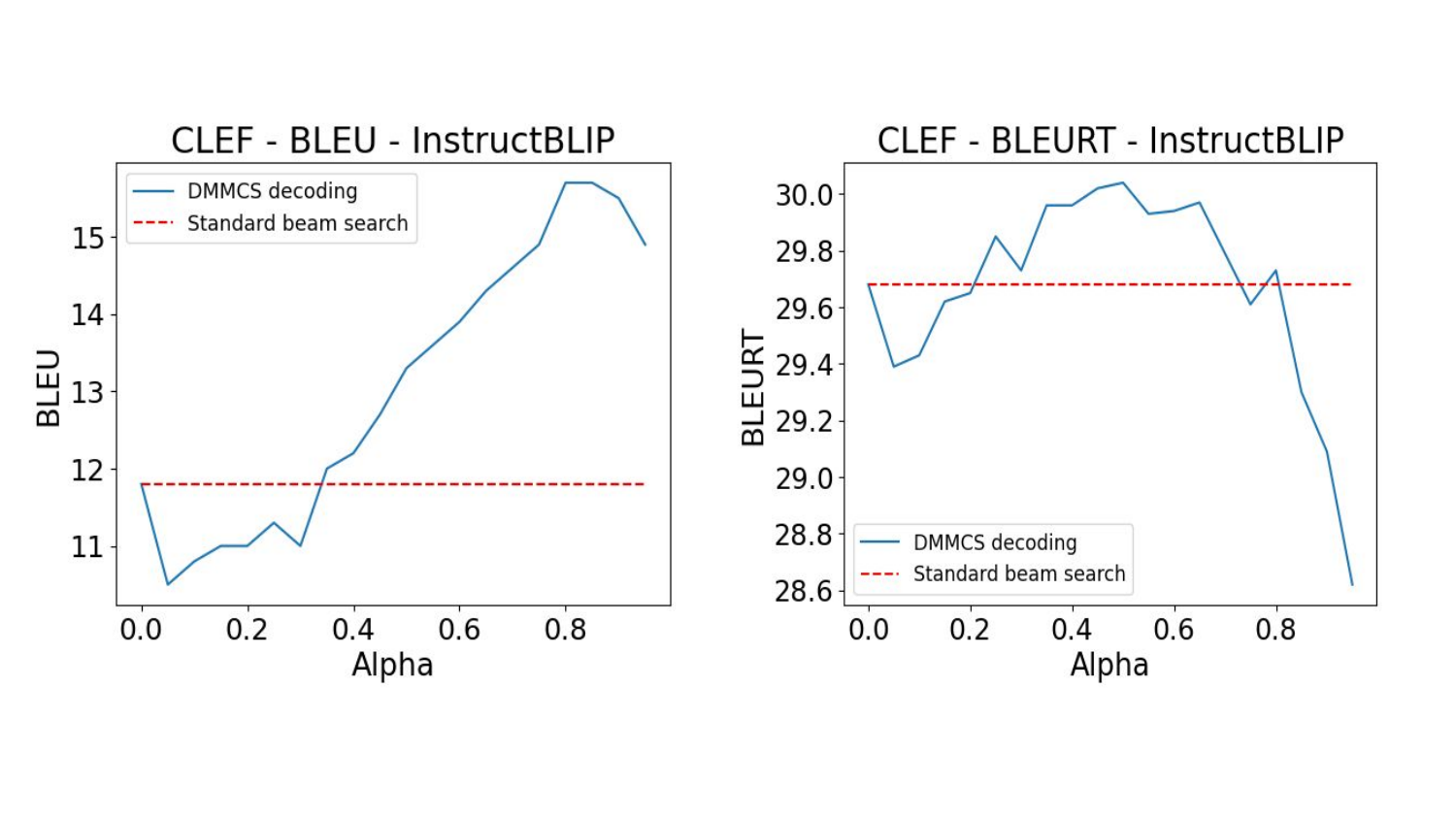}
    %\end{tabular}
\caption{InstructBLIP's performance with DMMCS decoding for various $\alpha$ values (horizontally) on ImageCLEF medical 2023, when using BLEU (left) or BLEURT (right) to measure performance. 
% dataset \cite{ImageCLEF2023}.
}
\label{fig:4.1}
\end{figure}

\smallskip\noindent\textbf{Tuning $\alpha$:} 
% The effect of the proposed penalty ($\textit{DMMCS}_{\textit{p}}$) and the decoder score for each candidate sequence ($\textit{D}_{\textit{score}}$) is regulated by the weighting factor $\alpha$, as shown in Eq.~\ref{eq6}. 
For each model, dataset, and evaluation measure, we tuned the $\alpha$ factor of DMMCS (Eq.~\ref{eq6}) by trying values from $0.05$ to $0.95$ and keeping the value with the best development score (measured with the particular evaluation measure).
% Thus, we performed a tuning process in order to decide the optimal value of $\alpha$ for each model. Specifically, we evaluated each model's performance on a held-out set for multiple values of $\alpha$, with an extensive search ranging from $0.05$ to $0.95$. 
As an example, Fig.~\ref{fig:4.1} shows the performance of InstructBLIP 
% \cite{instructblip}
on 
% the development subset of 
ImageCLEFmedical 2023, in terms of BLEU (left) or BLEURT (right), when using DMMCS (continuous lines) or standard beam search decoding (dotted horizontal lines), as a function of $\alpha$. For $\alpha$ values in $[0.4, 0.8]$, DMMCS improves the model's performance, while smaller and larger values deteriorate it, comparing to standard beam search.
% Yet, the performance of the model was enhanced with respect to all four metrics.

\begin{table*}[h!]
    \centering
    \begin{adjustbox}{max width=\textwidth}
    \begin{tabular}{|p{1.85cm}||p{1.28cm}|p{1.35cm}|p{1.35cm}|p{1.28cm}||p{1.28cm}|p{1.35cm}|p{1.35cm}|p{1.28cm}|}
        \hline
         \multicolumn{9}{|c|}{\textbf{ImageCLEFmedical 2023 Dataset - With Predicted Tags}} \\
        \hline
        \multicolumn{1}{|c||}{\multirow{4}{*}{}} & \multicolumn{4}{c||}{\footnotesize\textbf{BLEU}} & \multicolumn{4}{c|}{\footnotesize\textbf{BLEURT}} \\
        \cline{2-9}
        & \centering\footnotesize \textit{BS} & \centering\footnotesize \textit{ConBS $\forall$} & \centering\footnotesize \textit{ConBS $\exists$} & \centering\footnotesize \textit{DMMCS} & \centering\footnotesize \textit{BS} & \centering\footnotesize \textit{ConBS $\forall$} & \centering\footnotesize \textit{ConBS $\exists$} & \footnotesize \textit{DMMCS} \\
        \hline
        \footnotesize\centering \textbf{Show \& Tell} & \footnotesize \centering 20.61 \tiny (0.33)  & \footnotesize \centering 20.52 \tiny (0.39) & \footnotesize \centering 21.21 \tiny (0.38) & \footnotesize \centering \textbf{21.27} \tiny (0.35) & \footnotesize \centering 29.99 \tiny(0.14) & \footnotesize \centering 30.03 \tiny(0.17) & \footnotesize \centering 30.39 \tiny(0.13) &  \hspace{0.01cm}\footnotesize \textbf{30.47} \tiny(0.08)\\
        \hline
        \footnotesize\centering \textbf{ViT-GPT2} & \footnotesize \centering 15.34 \tiny(0.09) & \footnotesize \centering 15.75 \tiny(0.12) & \footnotesize \centering 16.29 \tiny(0.08) & \footnotesize \centering \textbf{16.31} \tiny(0.08) & \footnotesize \centering 26.50 \tiny(0.09) & \footnotesize \centering 26.31 \tiny(0.10) & \footnotesize \centering 26.92 \tiny(0.14) &  \hspace{0.01cm}\footnotesize \textbf{27.01} \tiny(0.16) \\
        \hline
        \footnotesize\centering \textbf{InstructBLIP} & \footnotesize \centering 11.81 \tiny(0.09) & \footnotesize \centering 15.89 \tiny (0.08) & \footnotesize \centering \textbf{16.14} \tiny (0.13) & \footnotesize \centering 15.93 \tiny(0.11) & \footnotesize \centering 29.68 \tiny(0.26) & \footnotesize \centering 29.71 \tiny(0.12) & \footnotesize \centering 30.08 \tiny(0.14) & \hspace{0.01cm}\footnotesize \textbf{30.10} \tiny(0.15) \\
        \hline
        \footnotesize\centering \textbf{Flamingo} & \footnotesize \centering 15.34 \tiny(0.11) & \footnotesize \centering 15.81 \tiny(0.13) & \footnotesize \centering \textbf{15.92} \tiny(0.06) & \footnotesize \centering 15.47 \tiny(0.09) & \footnotesize \centering 28.49 \tiny(0.19) & \footnotesize \centering 30.11 \tiny(0.21) & \footnotesize \centering 30.67 \tiny(0.19) &  \hspace{0.01cm}\footnotesize \textbf{31.34} \tiny(0.16) \\
        \hline
    \end{tabular}
    \end{adjustbox}
    \\
    \begin{adjustbox}{max width=\textwidth}
    \begin{tabular}{|p{1.85cm}||p{1.28cm}|p{1.35cm}|p{1.35cm}|p{1.28cm}||p{1.28cm}|p{1.35cm}|p{1.35cm}|p{1.28cm}|}
        \hline
         \multicolumn{9}{|c|}{\textbf{MIMIC-CXR Dataset - With Predicted Tags}} \\
        \hline
        \multicolumn{1}{|c||}{\multirow{4}{*}{}} & \multicolumn{4}{c||}{\footnotesize\textbf{BLEU}} & \multicolumn{4}{c|}{\footnotesize\textbf{BLEURT}} \\
        \cline{2-9}
        & \centering\footnotesize \textit{BS} & \centering\footnotesize \textit{ConBS $\forall$} & \centering\footnotesize \textit{ConBS $\exists$} & \centering\footnotesize \textit{DMMCS} & \centering\footnotesize \textit{BS} & \centering\footnotesize \textit{ConBS $\forall$} & \centering\footnotesize \textit{ConBS $\exists$} & \footnotesize \textit{DMMCS} \\
        \hline
        \centering \footnotesize\textbf{Show \& Tell} & \footnotesize \centering 11.77 \tiny(0.18) & \footnotesize \centering 12.14 \tiny(0.22) & \footnotesize \centering 13.38 \tiny(0.17) &  \footnotesize \centering \textbf{14.13} \tiny(0.24) & \footnotesize \centering 29.08 \tiny(0.27) & \footnotesize \centering 29.04 \tiny(0.29) & \footnotesize \centering 29.21 \tiny(0.31) & \hspace{0.01cm}\footnotesize \textbf{29.49} \tiny(0.29)\\
        \hline
        \centering \footnotesize\textbf{ViT-GPT2} & \footnotesize \centering 13.55 \tiny(0.26) & \footnotesize \centering 12.91 \tiny(0.19) & \footnotesize \centering 13.67 \tiny(0.23) & \footnotesize \centering \textbf{14.78} \tiny(0.21) & \footnotesize \centering 23.52 \tiny(0.34) & \footnotesize \centering 24.87 \tiny(0.37) & \footnotesize \centering \textbf{24.98} \tiny(0.30) & \hspace{0.01cm}\footnotesize 24.37 \tiny(0.29) \\
        \hline
        \centering \footnotesize \textbf{InstructBLIP} & \footnotesize \centering 12.76 \tiny(0.20) & \footnotesize \centering 12.09 \tiny(0.22) & \footnotesize \centering 12.27 \tiny(0.17) &  \footnotesize \centering \textbf{13.32} \tiny(0.19) & \footnotesize \centering 24.65 \tiny(0.24) & \footnotesize \centering 26.28 \tiny(0.27) & \footnotesize \centering \textbf{26.43} \tiny(0.26) & \hspace{0.01cm}\footnotesize 25.56 \tiny(0.25) \\
        \hline
        \centering \footnotesize \textbf{Flamingo} & \footnotesize \centering 12.78 \tiny(0.11) & \footnotesize \centering 13.07 \tiny(0.13) & \footnotesize \centering \textbf{13.39} \tiny(0.11) & \footnotesize \centering 13.26 \tiny(0.16) & \footnotesize \centering 29.14 \tiny(0.22) & \footnotesize \centering 28.86 \tiny(0.21) & \footnotesize \centering 29.58 \tiny(0.27) &  \hspace{0.01cm}\footnotesize \textbf{29.81} \tiny(0.24) \\
        \hline
    \end{tabular}
    \end{adjustbox}
    \\
    \begin{adjustbox}{max width=\textwidth}
    \begin{tabular}{|p{1.85cm}||p{1.28cm}|p{1.35cm}|p{1.35cm}|p{1.28cm}||p{1.28cm}|p{1.35cm}|p{1.35cm}|p{1.28cm}|}
    \hline
    \hline
    \centering \textit{\textbf{Wins}} & \centering\footnotesize 0 & \centering\footnotesize 0 & \centering\footnotesize 3 & \centering\footnotesize \textbf{5} & \centering\footnotesize 0 & \centering\footnotesize 0 & \centering\footnotesize 2 & \hspace{0.45cm}\footnotesize \textbf{6} \\
        \hline
    \end{tabular}
    \end{adjustbox}
    \caption{The performance of each model on both datasets, measured by BLEU or BLEURT. BS and ConBS denote beam search and constrained beam search decoding, respectively. $\forall$ and $\exists$ indicate whether ConBS is strict (all image tags must be expressed) or disjunctive (only one suffices), respectively. DMMCS is the new proposed decoding method. For the latter, tags predicted by a medical image tagger are used. We also report, along the evaluation metric score, the variance between the results of the three test subsets.}
    \label{fig:4.1.a}
    \vspace{0.2cm}
\end{table*}

\smallskip\noindent\textbf{BLEU, BLEURT results:} Table \ref{fig:4.1.a} reports the performance of each model on the two datasets, averaging over the three test subsets. For each evaluation measure (BLEU, BLEURT), four scores are provided: one for each baseline method (standard beam search, strict and disjunctive constrained beam search), as well as one for our proposed DMMCS method. The scores for DMMCS are obtained with the best $\alpha$ for each model, dataset, measure, using development data, as discussed above. Tags predicted by a medical image tagger are used. We observe that DMMCS always outperforms both standard (BS) and strict constrained ($\forall$) beam search. Moreover, DMMCS is on par with the disjunctive constrained ($\exists$) beam search method, outperforming it in most cases. We show the number of ``wins'' of each decoding method in the last row of Table \ref{fig:4.1.a}. We also  experimented using gold (oracle) tags instead of predicted tags; the results, discussed in  Appendix~\ref{sec:classifier}, show that DMMCS again improves performance, comparing to standard beam search.

%\vfill \null
\smallskip\noindent\textbf{Clinical accuracy results:} In Table~\ref{tab:clinical-accuracy}, we present the best performing method across models and datasets in terms of clinical accuracy (CA). When InstructBLIP is used as the backbone model, DMMCS is the best mechanism across datasets. For the rest of the models, however, there is no clear winner, although ConBS$\forall$ is the best in Mimic (see Appendix~\ref{sec:ca-results}). We  note that CA is based on silver labels (automatically generated), for a limited number of classes, and only on captions regarding chest (i.e., less than 20\%). Consequently, the reliability of these results may be compromised to some extent. Future work should focus on the development of better medical image taggers that could improve automated clinical evaluation (e.g., making it applicable not only to chest images).

\smallskip\noindent\textbf{Fluency results:}    In the last two rows of Table~\ref{tab:clinical-accuracy} we present the results based on fluency (using Perplexity, see Appendix~\ref{sec:perplexity-results}) instead of CA. Our proposed method is the best in both InstructBLIP and ViT-GPT2, and the best across models for ImageCLEFmedical 2023.

\begin{table*}[ht]
    \centering
    \begin{tabular}{rcccc}
               & \bf Show \& Tell & \bf ViT-GPT2 & \bf InstructBLIP & \bf Flamingo \\\hline
         \centering \bf ImageCLEFmedical 2023 & BS & ConBS $\exists$ & DMMCS & BS\\
         \centering \bf MIMIC-CXR & ConBS $\forall$ & ConBS $\forall$ & DMMCS & ConBS $\forall$\\\hline
         \centering \bf ImageCLEFmedical 2023 & DMMCS & DMMCS & DMMCS & DMMCS\\
         \centering \bf MIMIC-CXR & ConBS $\exists$ & DMMCS & DMMCS & BS\\
    \end{tabular}
    \caption{The first two rows present the most clinically accurate guided decoding method per model per dataset (more details in Appendix~\ref{sec:ca-results}). The last two rows present the same leaderboard, but use fluency (measured as Perplexity, Appendix~\ref{sec:perplexity-results}) instead of clinical accuracy.}
    \label{tab:clinical-accuracy}
\end{table*}

\begin{figure}[h!]
    %\centering
    \includegraphics[width=.48\textwidth]{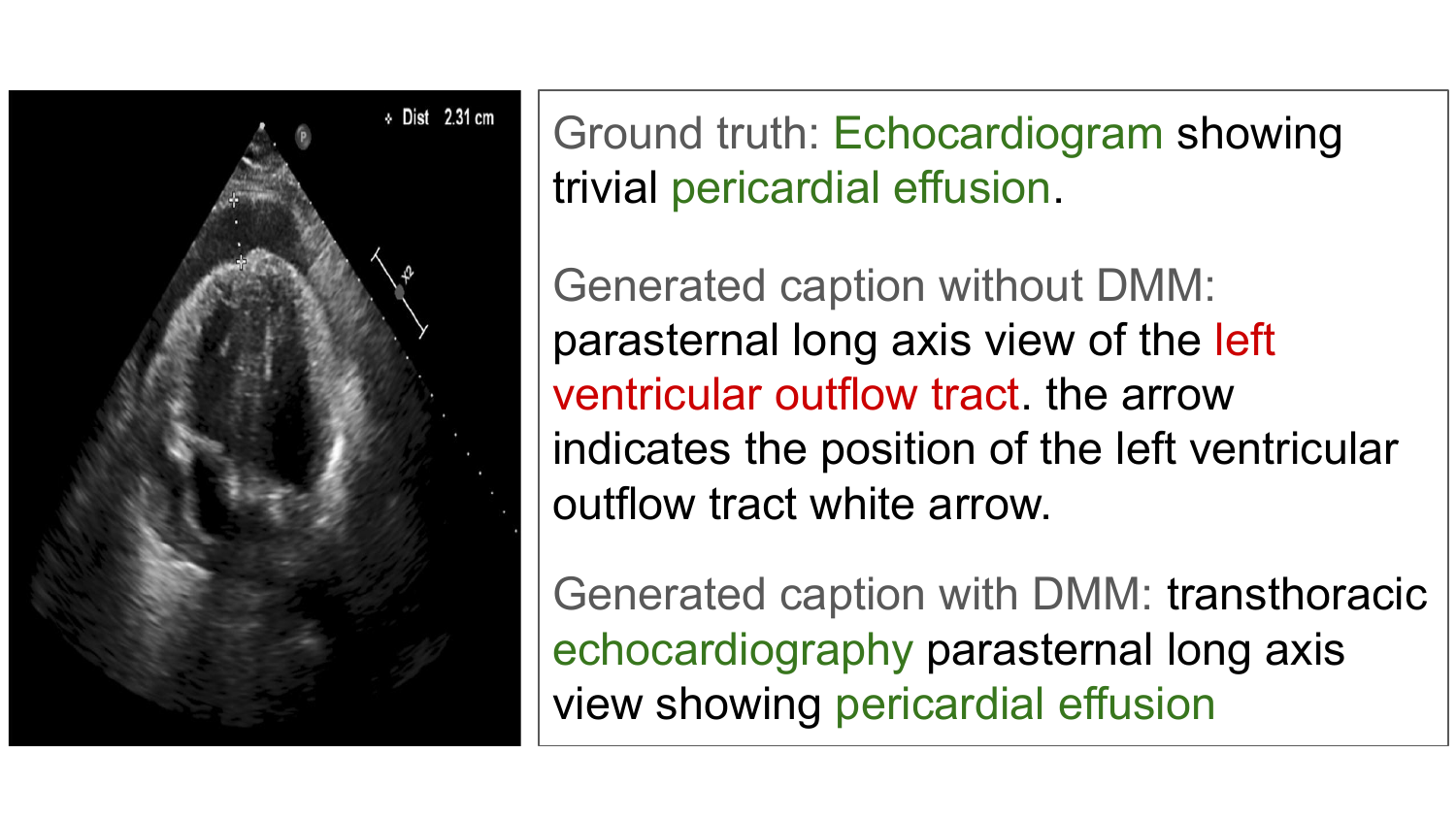}
    \vspace{-0.8cm}
\caption{A sample from an exploratory qualitative analysis of captions generated by ViT-GPT2, using standard beam search vs.\ DMMCS-enhanced decoding.}
\label{fig:4.3} 
\end{figure}

\smallskip\noindent\textbf{Qualitative results:} Additionally, we conducted a qualitative analysis of the diagnostic captions produced by the DC models with and without applying our proposed guided decoding method. We show a sample from this analysis in Figure \ref{fig:4.3}. Without DMMCS, the beam search decoder generated a partially inaccurate caption, referring to a ``parasternal long axis view'' instead of an ``echocardiogram''. While related, these terms are not precisely the same. Given the importance of precise medical terminology in diagnostic captions, addressing such inaccuracies is crucial. The DMMCS-enhanced caption rectified the examination type to an ``echocardiography parasternal long axis view'' aligning more closely with the reference caption. Moreover, it correctly identified the diagnosed medical condition as ``pericardial effusion'' without introducing extra inaccurate information, unlike beam search decoding which incorrectly stated that the examination type showed a ``left ventricular outflow tract''. However, it is crucial to note that the proposed method may not address all hallucinations, or more generally, inaccuracies of the model. Nevertheless, the draft reports generated by our method are more accurate according to all evaluation metrics, and in practice they would be checked and improved by medical experts.
% Nevertheless, the draft reports generated by our method are more accurate according to all evaluation metrics. The reader is also reminded that we aim to generate draft captions that will be checked and improved by medical experts.

\smallskip\noindent\textbf{Varying Sentence Order Analysis:} The sentence order in medical reports varies (e.g., the same diagnosis may appear in different reports, using the same sentences yet reordered). In light of this observation, we conducted a sentence-level analysis to investigate the potential influence of sentence order on the quality of the generated captions. We considered pairs of gold and generated captions consisting of the same number of sentences. We then created sentence pairs using sentences at the same position in the gold and generated captions. We evaluated each pair in terms of BLEU and BLEURT and then averaged the sentence-pair scores across the test set. If the order of the generated sentences is often wrong compared to the corresponding gold ones, we should notice a substantial difference relative to the original scores, which do not penalize generating reasonably correct sentences but in the wrong order. Our analysis, however, revealed that the scores of sentence pairs remained consistent with the original ones (within the reported standard deviation), ultimately indicating that this issue did not \linebreak affect our experiments.

\smallskip\noindent\textbf{Cases where performance declines:} As shown in Tables \ref{fig:4.1.a} and \ref{tab:clinical-accuracy}, the proposed algorithm generally improves performance on both Natural Language Generation (NLG) metrics and clinical accuracy. However, there are instances where the performance declines when using the proposed algorithm. We investigated these cases by manually examining them, but no specific pattern explaining the decrease in performance could be identified. We plan to investigate this issue further in the future, possibly requiring assistance from medical professionals.

\smallskip\noindent \textbf{Computational Overhead:} It is important to measure the computational overhead associated with our proposed method. Implementing our method results in an additional time overhead, requiring approximately $25$ to $27\%$ more time to generate diagnostic captions compared to baseline methods such as standard beam search. In the case of InstructBLIP, this translates to an increase of around $13$ to $15$ minutes in processing time per $1,000$ captions in our experiments. Furthermore, there is an additional computational overhead in terms of memory usage, which increases by approximately $5\%$ compared to baseline approaches (Section~\ref{sec:baselines}). While these overheads are important considerations, the improved performance, in terms of NLG metrics and clinical accuracy, of the generated captions justify the investment in time and computational resources.

\smallskip\noindent \textbf{Per-Modality Results} The ImageCLEFmedical 2023 dataset consists of four primary medical modalities; namely X-Ray, CT, MRI and Ultrasonography. We explored the individual per-modality results of both the standard beam search and the proposed decoding method in order to identify any modality-specific features that might require attention. We observed that the proposed algorithm consistently outperforms the standard beam search across both metrics and nearly all modalities and models. No modality-specific peculiarities were noted, so we did not deem it necessary to implement any case-specific data handling procedures. Detailed per-modality evaluations are provided in Table \ref{fig:per-modality-scores} (Appendix \ref{per-modality}).

\section{Discussion}

\subsection*{Performance with gold tags}\label{ssec:gold-tags}
% Theoretical upper bound estimation
% To approximate the performance of the four models on both datasets, 
We also experimented with the ground-truth (instead of the predicted) tags per image (see also Table~\ref{fig:4.2} in the Appendix). As one would expect, DMMCS-enhanced models perform better with gold, compared to predicted tags. 
% DMMCS enhances model performance compared to standard beam search and results in superior model performance. The extent of this improvement depends on the accuracy of the classifier, with less accurate models generating noisier tags and, subsequently, affecting the generated captions. Furthermore, 
However, the performance of DMMCS-enhanced models with predicted tags (Table~\ref{fig:4.1.a}) is relatively close to the corresponding performance with ground-truth tags (Table~\ref{fig:4.2}), showing the robustness of DMMCS to noisy tags. 
% Hence, despite the use of noisy predicted tags in real-life settings, the model's performance remained competitive, underscoring the effectiveness of the proposed decoding algorithm.

\subsection*{Guidance balancing}\label{ssec:histogram-divergence}
We also experimented by balancing the two basic components of DMMCS (Eq.\ \ref{eq6}), the decoder's scores and the proposed data-driven penalty. By introducing a dynamically computed weight called Histogram Divergence (HD), we adjusted the contribution of the two components, so that when DMMCS could not be trusted, more weight would be assigned to the decoder's score  (more details in Appendix \ref{apx:histogram}). We observe that this balancing results in steadier performance across $\alpha$ values, though it comes at the cost of lower overall performance compared to when HD is not considered.
In other words, if tuning is not an option, dynamically balancing the two components of DMMCS could lead to the selection of a better optimal $\alpha$.

\section{Conclusion}
In this work we introduced the DMMCS data-driven guided decoding method that enhances the performance of automatically generated diagnostic captions by integrating medical tags associated with a radiology image in the generation process. We assessed the effectiveness of the proposed method by applying it to four DC systems with diverse architectures and on two datasets, namely ImageCLEFmedical 2023 and a subset of MIMIC-CXR. Subsequently, we compared the generated captions with those obtained using a typical beam search decoding approach, employing widely used evaluation metrics, namely BLEU and BLEURT. Our results demonstrate that using the proposed DMMCS mechanism during decoding consistently outperforms the typical beam search approach across almost all models and datasets for most of the metrics.

In future work, we plan to experiment with more domains and focus on a broader range of tasks to investigate the benefits of our method in a wider context. Furthermore, we will explore our method's capabilities in generic image captioning \cite{lin2015microsoft}, as well as other text generation tasks. %including question-answering, text summarization, and machine translation. 
A final direction for future work concerns the use of contextual representations, in order to enhance the quality of the embeddings used when computing the penalty of DMMCS (Eq.~\ref{eq5}).

% In future work, we plan to experiment with a broader range of datasets and tasks. Furthermore, we aim to explore our method's capabilities in generic (non-medical) image captioning, as well as other text generation tasks, including question-answering, text summarization, and machine translation. Regarding biomedical question answering, we plan to investigate whether our decoding method can be applied based on snippets provided by medical experts \cite{10.1007/978-3-031-42448-9_19} or extracted using Retrieval Augmented Generation \cite{10.5555/3495724.3496517}. Finally, we intend to conduct a series of ablation studies in order to analyze the contribution and impact of the algorithm's different components, as well as to what extent the tags' $\textit{MCS}$ distributions provide interpretability of the generated captions with respect to the concepts.

%\clearpage
\section{Limitations}
\label{sec:limitations}

% The generalisability of the proposed method is limited by the constrained number of datasets used in this study. 
Although we used all the publicly available medical datasets we could obtain, the experimental results are limited to the specific conditions, regions, and language (English) these datasets concern. However, this limitation could be addressed by collaborating with medical institutions, under the license of respective review boards, which is a direction we plan to consider in the future.
    %LLMs present several limitations including hallucinations (making up facts that are not true about the world), difficulty to update the world knowledge stored in their parameters, as well as to access and apply that knowledge for a particular task, which are potentially inherited in our method when employing these models as well.
    % \item The success of the proposed method depends on the quality of the medical tags. This means that noisy input (tags) could affect the performance. To investigate this hypothesis, besides the gold tags, we also used tags retrieved by an image classifier for the ImageCLEF medical 2023 dataset. As it is illustrated at the top rows of Table \ref{fig:4.2} (\S\ref{sec:results}), the increase in performance with \textit{DMMCS}-based decoding when using the predicted tags slightly drops.
    %\item Clinical correctness of automatically-generated diagnostic reports is critical. Although our \textit{DMMCS}-based generated reports were more accurate than the baseline, we note that their clinical correctness has not been assessed. That is mainly due to the absence of appropriate evaluation measures, which is a direction that future research should focus on.

\section{Ethical Considerations}
% Our proposed \textit{DMMCS}-based decoding method is related to the $3^{\text{rd}}$ and the $10^{\text{th}}$ goal of United Nations Sustainability Goals, since a
Assisting clinicians by providing them more accurate draft diagnostic reports promotes ensuring good health and well-being, as well as reduced inequalities. However, emphasis on biomedical data privacy has long been a sensitive issue because of crossing ethical, legal, and technical boundaries, thus apprehension of clinical information privacy needs to be taken into serious consideration when patients' data are used for model training.

% Moreover, when ground-truth tags are not provided and need to be retrieved from a patient's medical history, developed countries, where medical information is typically stored in large information systems, have a clear advantage over undeveloped regions or countries under development. In these situations, a medical image classifier can be employed to predict the medical concepts depicted in an image, as it is demonstrated in Appendix \ref{sec:classifier}. We believe that the positive societal impact associated to using our method exceeds potential problems and limitations of DC systems in general.
%\end{itemize}

\section*{Acknowledgements}
This work has been partially supported by project MIS 5154714 of the National Recovery and Resilience Plan Greece 2.0 funded by the European Union under the NextGenerationEU Program.

\bibliography{anthology,custom}

\section*{Appendix}
\appendix
\section{Background on DC}
\label{sec:dc-background}

Various DC approaches have been explored in previous work, ranging from earlier ontology and rule-based systems \cite{Varges2012SemScribeNL} to modern encoder-decoder architectures. Notably, the best-performing techniques in generic image captioning \cite{generic-image-captioning-survey} are not always the most efficient for DC tasks. Unlike generic image captioning models, DC systems face challenges in accurately describing medical images based solely on visual content. For instance, conventional retrieval methods, despite their simplicity, have shown promising results in DC tasks by leveraging captions from similar archived exams \cite{Pavlopoulos2021DiagnosticCA}. Even simple models like $1$-NN, which retrieve and use the caption from the visually nearest image, can outperform more complex systems, such as Transformer-based or encoder-decoder architectures \cite{pmlr-v106-liu19a}. However, despite the noteworthy performance of retrieval methods, recent advancements in the field of deep learning have established the encoder-decoder framework as the predominant approach in the DC domain. CNNs or Vision Transformers \cite{vit} are commonly employed for image encoding, while RNNs or Transformer-based LMs are selected for the report generation process. Recent advancements in diagnostic captioning have also incorporated attention mechanisms, visual attention, and reinforcement learning techniques into the encoder-decoder framework \cite{dc-reinforcement}.

% \noindent 
%The model proposed by \cite{yang-etal-dc} currently stands as a competitive diagnostic captioning system for chest radiology data. It integrates general and task-specific knowledge to improve medical report quality. The architecture comprises four modules: general knowledge embedding, visual feature extraction, specific knowledge retrieval and embedding, and report generator. RadGraph \cite{radgraph}, a manually built knowledge graph, provides general medical knowledge, which is combined with input visual features through multi-head attention. Task-specific knowledge is retrieved from a medical database and incorporated using ClinicalBERT \cite{clinicalbert}. The fusion of these components enhances report generation, achieved through multiple attention and feed-forward layers.

\begin{table*}[h!]
    \centering
    \footnotesize
    \begin{tabular}{|p{2.15cm}||p{1cm}|p{1cm}||p{1cm}|p{1cm}||p{1cm}|p{1cm}||p{1cm}|p{1cm}|p{1cm}|}
        \hline
         \multicolumn{7}{|c|}{\textbf{A. ImageCLEFmedical 2023 Dataset - Ground-Truth Tags}} \\
        \hline
        \multicolumn{1}{|c||}{\multirow{4}{*}{}} & \multicolumn{2}{c||}{\footnotesize\textbf{BLEU}} & \multicolumn{2}{c||}{\textbf{ROUGE}} & \multicolumn{2}{c|}{\textbf{BLEURT}} \\
        \cline{2-7}
        & \centering\footnotesize \textit{BS} & \centering\footnotesize \textit{DMMCS} & \centering\footnotesize \textit{BS} & \centering\footnotesize \textit{DMMCS} & \centering\footnotesize \textit{BS} & \footnotesize \textit{DMMCS} \\
        \hline
        \centering \footnotesize\textbf{Show \& Tell} & \footnotesize \centering 20.61 & \footnotesize \centering \textbf{21.48} & \footnotesize \centering 22.00 & \footnotesize \centering \textbf{23.01} & \footnotesize \centering 29.99 & \hspace{0.2cm}\footnotesize \textbf{30.53}\\
        \hline
        \centering \footnotesize\textbf{ViT-GPT2} & \footnotesize \centering 15.34 & \footnotesize \centering \textbf{16.72} & \footnotesize \centering 16.72 & \footnotesize \centering \textbf{17.05} & \footnotesize \centering 26.50 & \hspace{0.2cm}\footnotesize \textbf{27.69} \\
        \hline
        \centering \textbf{InstructBLIP} & \footnotesize \centering 11.81 & \footnotesize \centering \textbf{19.95} & \footnotesize \centering 20.98 & \footnotesize \centering \textbf{21.17} & \footnotesize \centering 29.68 & \hspace{0.2cm}\footnotesize \textbf{30.65} \\
        \hline
        \centering \textbf{Flamingo} & \footnotesize \centering 15.34 & \footnotesize \centering \textbf{15.83} & \footnotesize \centering 15.98 & \footnotesize \centering \textbf{16.11} & \footnotesize \centering 28.49 & \hspace{0.2cm}\footnotesize \textbf{31.47} \\
        \hline
    \end{tabular}
    \\
    \begin{tabular}{|p{2.15cm}||p{1cm}|p{1cm}||p{1cm}|p{1cm}||p{1cm}|p{1cm}||p{1cm}|p{1cm}|p{1cm}|}
        \hline
         \multicolumn{7}{|c|}{\textbf{B. MIMIC-CXR Dataset - Ground-Truth Tags}} \\
        \hline
        \multicolumn{1}{|c||}{\multirow{4}{*}{}} & \multicolumn{2}{c||}{\textbf{BLEU}} & \multicolumn{2}{c||}{\textbf{ROUGE}} & \multicolumn{2}{c|}{\textbf{BLEURT}} \\
        \cline{2-7}
        & \centering\footnotesize \textit{BS} & \centering\footnotesize \textit{DMMCS} & \centering\footnotesize \textit{BS} & \centering\footnotesize \textit{DMMCS} & \centering\footnotesize \textit{BS} & \footnotesize \textit{DMMCS} \\
        \hline
        \centering \textbf{Show \& Tell} & \footnotesize \centering 11.77 & \footnotesize \centering \textbf{14.29} & \footnotesize \centering \textbf{17.18} & \footnotesize \centering 17.10 & \footnotesize \centering 29.08 & \hspace{0.2cm}\footnotesize \textbf{29.65}\\
        \hline
        \centering \textbf{ViT-GPT2} & \footnotesize \centering 13.55 & \footnotesize \centering \textbf{15.01} & \footnotesize \centering 21.37 & \footnotesize \centering \textbf{21.74} & \footnotesize \centering 23.52 & \hspace{0.2cm}\footnotesize \textbf{24.43} \\
        \hline
        \centering \textbf{InstructBLIP} & \footnotesize \centering 12.76 & \footnotesize \centering \textbf{13.39} & \footnotesize \centering 14.35 & \footnotesize \centering \textbf{16.27} & \footnotesize \centering 24.65 & \hspace{0.2cm}\footnotesize \textbf{25.61} \\
        \hline
        \centering \textbf{Flamingo} & \footnotesize \centering 12.78 & \footnotesize \centering \textbf{13.99} & \footnotesize \centering 13.68 & \footnotesize \centering \textbf{13.86} & \footnotesize \centering 29.14 & \hspace{0.2cm}\footnotesize \textbf{29.87} \\
        \hline
    \end{tabular}
    \caption{The performance of each model on both datasets evaluated across the three metrics. \textit{BS} and \textit{DMMCS} denote beam search and DMMCS-based decoding respectively. For the DMMCS-based decoding, ground-truth tags provided in the training set are used.}
    \label{fig:4.2}
\end{table*}

\section{DMMCS based on ground-truth tags}
\label{sec:classifier}
In this section, we report the performance of the four models on both datasets using the \emph{ground-truth} tags for each image (Table~\ref{fig:4.2}), instead of predicted tags (Table~\ref{fig:4.1.a}). 
As shown in the main section of the paper, employing the proposed algorithm enhances model performance compared to standard beam search. Furthermore, employing ground-truth tags, instead of tags predicted by a medical image classifier, results in superior model performance. The extent of this performance difference depends on the accuracy of the classifier. 
% As discussed in \S\ref{sec:limitations}, 
Less accurate classifiers generate noisier tags, which subsequently affect the generated captions. Conversely, more accurate classifiers provide guidance for the model to include relevant concepts, as well as words closely aligned with the medical examination's findings. In our experiments, we observed that the performance of the models utilizing predicted tags \linebreak (Table \ref{fig:4.1.a}) was relatively close to the ones employing ground-truth tags (Table \ref{fig:4.2}). This indicates that despite the use of predicted tags, the model's performance remained competitive, indicating the robustness of the proposed decoding algorithm. 

\section{ROUGE scores}
\label{sec:rouge}
In the main part of this paper, we primarily focused on a precision-based measure (i.e., BLEU \cite{papineni-etal-2002-bleu2}) and a learned evaluation metric (i.e., BLEURT \cite{sellam-etal-2020-bleurt2}) that relies on precision-driven metrics during training. BLEU evaluates the precision of n-grams in the generated text compared to reference texts, while BLEURT is a learned metric that leverages transformer-based embeddings and is trained on human judgment data. It incorporates BLEU in order to enhance its evaluation capabilities.
Nevertheless, recall is also important, as it measures the model's ability to capture all relevant details from the reference captions. Unlike precision, which emphasizes the correctness of the generated information, recall ensures that the model doesn't miss any pertinent details. However, clinicians face time constraints and cannot feasibly process overly long or potentially inaccurate information. Therefore, while recall is crucial for ensuring thoroughness, it must be balanced with the practical considerations of clinical workflow. Our main focus remains on precision, ensuring that the information provided is concise, relevant, and accurate. Nevertheless, the model's performance in a recall-based metric, specifically ROUGE \cite{lin-2004-rouge2}, is also included in this section (Table \ref{fig:4.2}). This approach allows us to provide a more comprehensive evaluation while maintaining a focus on the precision needed for practical clinical use.

\begin{figure}[h!]
\vspace{0.2cm}
    \centering
    \includegraphics[width=8.5cm, height=7.6cm, trim={2.3cm 0cm 0cm 3.5cm}, clip]{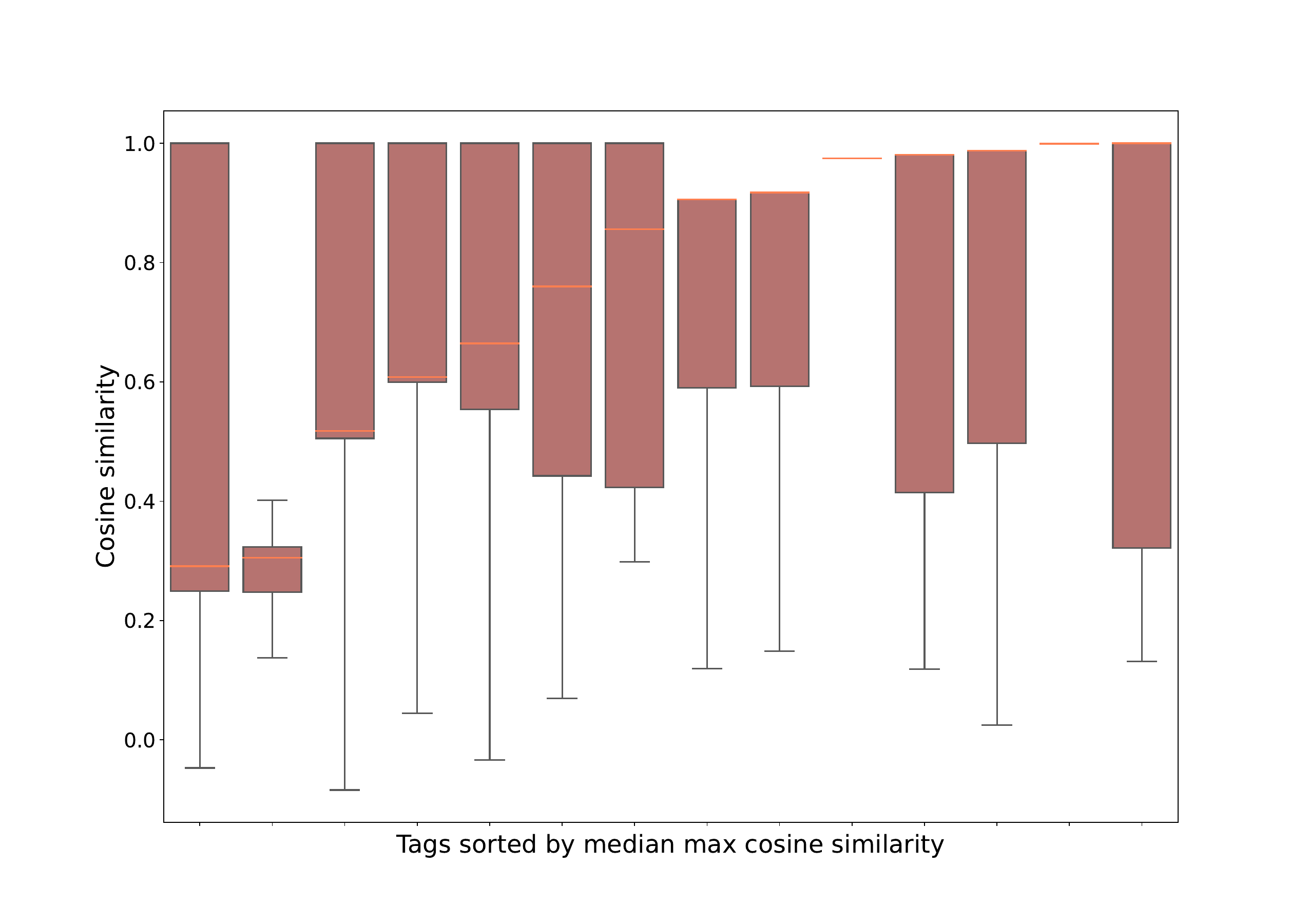}
    \vspace{-0.8cm}
\caption{ \label{fig:mimic3.3} Multiple boxplots plotted side-by-side. Each plot visualizes the $\textit{MCS}$ distribution between a tag and its associated training captions in the MIMIC-CXR dataset. The boxplots are sorted based on their median value (coral line), which denotes the tag’s median maximum cosine similarity (MMCS) value.}
\end{figure}

\section{MIMIC-CXR MCS distributions}
In this part, we also provide a figure with multiple interquartile plots for the employed subset of the MIMIC-CXR \cite{mimic-cxr} dataset, in correspondance with Figure~\ref{fig:3.3}. Each plot represents the MCS distribution of each one of the $14$ tags. The coral lines denote the median value of each interquartile plot (MMCS). We observe that, similarly to the ImageCLEFmedical 2023 dataset (see Figure \ref{fig:3.3}), the MMCS value of the distinct tags vary significantly, ranging from $0.3$ to $1.0$.

\section{Histogram Divergence (HD)}\label{apx:histogram}
As part of further exploration, we experimented with adding an additional term in our method's guided decoding scoring function. Histogram Divergence (HD) is a dynamically computed parameter that serves as a weighting factor of the two components, $\textit{DMMCS}_{\textit{p}}$ and $1-\textit{D}_{\textit{score}}$. In detail, at each decoding step, we consider a set of generated candidate sequences $G$ and a histogram of the MCS distribution $R(t, G)$ (Eq.~\ref{eq3}) for each tag $t$ is calculated. It represents the maximum lexical representation of the tag $t$ in each candidate sequence. Therefore, for a given list of ground-truth tags $T$, $|T|$ histograms are
generated. In addition, a similar histogram has been pre-calculated for each tag on its associated ground-truth captions with respect to the training data (see Section ~\ref{sec:method}). The rationale behind HD is that in
candidate sequences with similar \textit{MCS} distributions (one for each tag)  to those computed on the training data the $\textit{DMMCS}_{\textit{p}}$ factor should be assigned a larger weight compared to $1-\textit{D}_{\textit{score}}$ during that decoding step. In contrast, if the distributions do not match, the conventional $\textit{D}_{\textit{score}}$ should be trusted more.

Given a single tag $c$, the divergence of its two corresponding histograms (computed on the generated and ground-truth captions respectively) is calculated by performing a Kolmogorov-Smirnov Goodness of Fit Test (KS-test) \cite{kstest}, which yields two values: the p-value and the ks-statistic. The p-value, as in any other goodness of fit test, can be examined to either accept or reject a null hypothesis, but it does not precisely quantify the discrepancy between the two provided distributions. Unlike p-value, the ks-statistic measures the distance between two given distributions, while it is an appropriate metric in the specified context since it takes values in $[0,1]$. A value of $0$ suggests that the two samples are drawn from the same distribution, while a value of $1$ indicates the opposite. The ks-statistic is calculated as the maximum absolute vertical distance between the Empirical Cumulative Distribution Functions (ECDF) of the two distributions. Formally, the ks-statistic for a single tag $t$ at a random decoding step can be calculated as:

\begin{align*}
\label{eq8}
\tag{8}
   \textit{ks}(t) = \max |F(R(t, S)) - F(R(t, G))|,
\end{align*}

\noindent where $S$ denotes the training captions associated with $t$ and $G$ represents the generated captions up to this decoding step. Moreover, $F(\cdot)$ calculates the ECDF of the given input, while $R$ is the \textit{MCS} distribution as defined in Eq.~\ref{eq3}. Overall, the HD for a set of tags $T$ is calculated as:

\begin{align*}
\label{eq9}
\tag{9}
    %\textit{HD} = \sum_{t \in T} \frac{\textit{ks}(t)}{|T|},
    \textit{HD} = \frac{1}{|T|} \cdot \sum_{t \in T} \textit{ks}(t).
\end{align*}

\noindent HD is integrated in Eq.~\ref{eq6} as an additional weighting factor:

\begin{align*}
    \textit{DMMCS}(s) = & \ \alpha \cdot (1 - \textit{HD}) \cdot \textit{DMMCS}_{\textit{p}}(T,s) \\ 
    & + (1-\alpha) \cdot \textit{HD} \cdot (1-\textit{D}_{\textit{score}})
\label{eq10}
\tag{10}
\end{align*}

\noindent As can be seen in Fig.~\ref{fig:mimic3.3}, both blue (with HD) and green (without HD) lines surpass the red dashed line of the beam search baseline for various $\alpha$ values when using the (oracle) gold tags. The green line achieves a higher peak compared to the blue, while the blue line has lower standard deviation. Therefore, HD improves robustness across $\alpha$ values at the cost of (maximum) performance. 

\begin{figure}[ht]
    \centering
    \includegraphics[width=\linewidth,trim={0cm 0.5cm 1cm 0cm},clip]{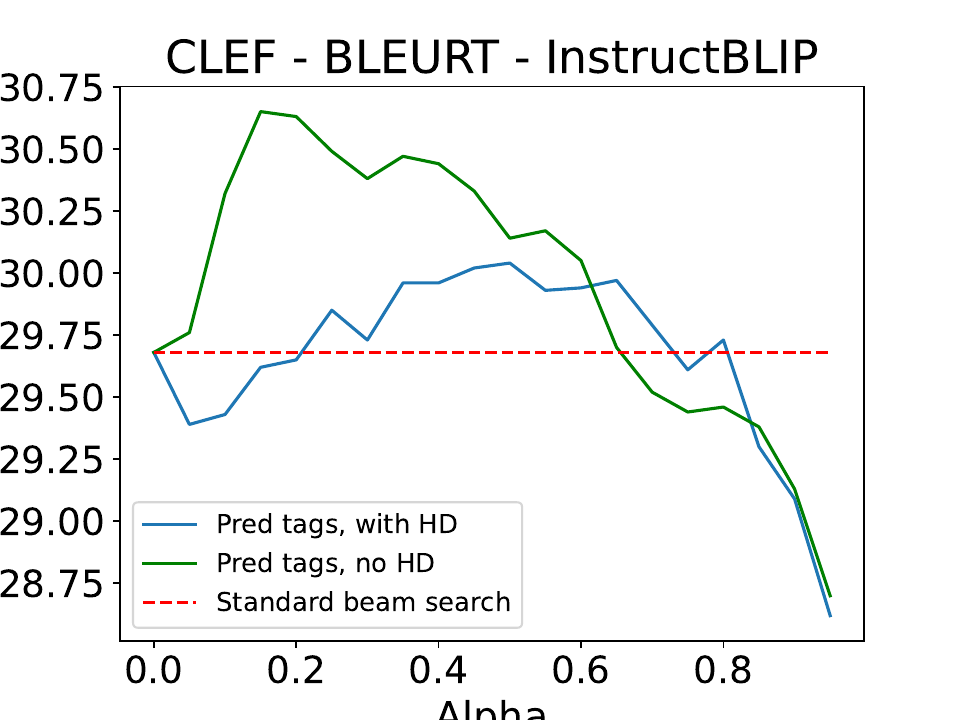}
\caption{InstructBLIP's performance in BLEURT with (blue) and without (green) HD in ImageCLEFmedical 2023 for various $\alpha$ values (horizontally). The red line denotes the performance of the standard beam search baseline.}
\label{fig:mimic3.3}
\end{figure}
%\noindent An illustrated example aiming to assist with the comprehension of the ks-statistic, and equivalently the calculation of the HD weight, is provided in Appendix A.
%\\

%At a decoding step $n$, the LM score for a given candidate caption $s$ is calculated as:

%\begin{align*}
%\label{eq9}
%\tag{9}
%    \textit{LM}_{\textit{score}} = - \sum_{n=1}^{|s|} \log %P(s_n | s_{<n}).
%\end{align*}

%\noindent It is normalized between $0$ and $1$ in order to align with the $\textit{DMM}_{\textit{p}}$ score's range. As the goal is to minimize the overall DMM score, the mathematical formula in Equation \ref{eq6} incorporates the normalized $\textit{LM}_{\textit{score}}$'s distance from 1.

\begin{table*}[h!]
    \centering
    \begin{adjustbox}{max width=\textwidth}
    \begin{tabular}{|p{1.85cm}||p{1.28cm}|p{1.35cm}||p{1.35cm}|p{1.28cm}||p{1.28cm}|p{1.35cm}||p{1.35cm}|p{1.28cm}|}
        \hline
         \multicolumn{9}{|c|}{\textbf{ImageCLEFmedical 2023 Dataset - Per modality Evaluation - With Predicted Tags}} \\
        \hline
        \multicolumn{1}{|c||}{\multirow{4}{*}{}} & \multicolumn{8}{c|}{\footnotesize\textbf{BLEU}} \\
        \cline{2-9}
        \multicolumn{1}{|c||}{\multirow{4}{*}{}} &
        \multicolumn{2}{c||}{\footnotesize\textbf{X-Ray}}&
        \multicolumn{2}{c||}{\footnotesize\textbf{CT}}&
        \multicolumn{2}{c||}{\footnotesize\textbf{MRI}}&
        \multicolumn{2}{c|}{\footnotesize\textbf{Ultrasonography}}\\
        \hline
        & \centering\footnotesize \textit{BS} & \centering\footnotesize \textit{DMMCS} & \centering\footnotesize \textit{BS} & \centering\footnotesize \textit{DMMCS} & \centering\footnotesize \textit{BS} & \centering\footnotesize \textit{DMMCS} & \centering\footnotesize \textit{BS} & \footnotesize \textit{DMMCS} \\
        \hline
        \footnotesize\centering \textbf{Show \& Tell} & \footnotesize \centering 20.73  & \footnotesize \centering \textbf{21.52} & \footnotesize \centering 20.71 & \footnotesize \centering \textbf{21.29} & \footnotesize \centering 20.54 & \footnotesize \centering \textbf{21.21} & \footnotesize \centering 20.46 &  \hspace{0.25cm}\footnotesize \textbf{21.06} \\
        \hline
        \footnotesize\centering \textbf{ViT-GPT2} & \footnotesize \centering 15.44 & \footnotesize \centering \textbf{16.62} & \footnotesize \centering 15.58 & \footnotesize \centering \textbf{16.21} & \footnotesize \centering 15.49 & \footnotesize \centering \textbf{15.87} & \footnotesize \centering 14.85 &  \hspace{0.25cm}\footnotesize \textbf{16.54} \\
        \hline
        \footnotesize\centering \textbf{InstructBLIP} & \footnotesize \centering 11.72 & \footnotesize \centering \textbf{15.92} & \footnotesize \centering 11.94 & \footnotesize \centering \textbf{16.07} & \footnotesize \centering 11.80 & \footnotesize \centering \textbf{16.01} & \footnotesize \centering 11.79 & \hspace{0.25cm}\footnotesize \textbf{15.72} \\
        \hline
        \footnotesize\centering \textbf{Flamingo} & \footnotesize \centering 15.39 & \footnotesize \centering \textbf{15.63} & \footnotesize \centering 15.23 & \footnotesize \centering \textbf{15.67} & \footnotesize \centering \textbf{15.41} & \footnotesize \centering 15.36 & \footnotesize \centering \textbf{15.25} &  \hspace{0.25cm}\footnotesize 15.22 \\
        \hline
    \end{tabular}
    \end{adjustbox}
    \\
    \begin{adjustbox}{max width=\textwidth}
    \begin{tabular}{|p{1.85cm}||p{1.28cm}|p{1.35cm}||p{1.35cm}|p{1.28cm}||p{1.28cm}|p{1.35cm}||p{1.35cm}|p{1.28cm}|}
        \hline
        \multicolumn{1}{|c||}{\multirow{4}{*}{}} & \multicolumn{8}{c|}{\footnotesize\textbf{BLEURT}} \\
        \cline{2-9}
        \multicolumn{1}{|c||}{\multirow{4}{*}{}} &
        \multicolumn{2}{c||}{\footnotesize\textbf{X-Ray}}&
        \multicolumn{2}{c||}{\footnotesize\textbf{CT}}&
        \multicolumn{2}{c||}{\footnotesize\textbf{MRI}}&
        \multicolumn{2}{c|}{\footnotesize\textbf{Ultrasonography}}\\
        \hline
        & \centering\footnotesize \textit{BS} & \centering\footnotesize \textit{DMMCS} & \centering\footnotesize \textit{BS} & \centering\footnotesize \textit{DMMCS} & \centering\footnotesize \textit{BS} & \centering\footnotesize \textit{DMMCS} & \centering\footnotesize \textit{BS} & \footnotesize \textit{DMMCS} \\
        \hline
        \footnotesize\centering \textbf{Show \& Tell} & \footnotesize \centering 30.13  & \footnotesize \centering \textbf{30.39} & \footnotesize \centering 29.96 & \footnotesize \centering \textbf{30.52} & \footnotesize \centering 30.03 & \footnotesize \centering \textbf{30.42} & \footnotesize \centering 29.84 &  \hspace{0.25cm}\footnotesize \textbf{30.55} \\
        \hline
        \footnotesize\centering \textbf{ViT-GPT2} & \footnotesize \centering 26.42 & \footnotesize \centering \textbf{27.17} & \footnotesize \centering 26.71 & \footnotesize \centering \textbf{26.82} & \footnotesize \centering 26.36 & \footnotesize \centering \textbf{26.94} & \footnotesize \centering 26.51 &  \hspace{0.25cm}\footnotesize \textbf{27.11} \\
        \hline
        \footnotesize\centering \textbf{InstructBLIP} & \footnotesize \centering 29.83 & \footnotesize \centering \textbf{30.13} & \footnotesize \centering 29.45& \footnotesize \centering \textbf{30.01} & \footnotesize \centering 29.70 & \footnotesize \centering \textbf{30.27} & \footnotesize \centering 29.74 & \hspace{0.25cm}\footnotesize \textbf{29.99} \\
        \hline
        \footnotesize\centering \textbf{Flamingo} & \footnotesize \centering 28.64 & \footnotesize \centering \textbf{31.33} & \footnotesize \centering 28.52 & \footnotesize \centering \textbf{31.61} & \footnotesize \centering 28.03 & \footnotesize \centering \textbf{31.31} & \footnotesize \centering 28.77 &  \hspace{0.25cm}\footnotesize \textbf{31.11} \\
        \hline
    \end{tabular}
    \end{adjustbox}
    \caption{The per-modality performance of each model on the ImageCLEFmedical 2023 dataset, measured by BLEU and BLEURT. BS denotes beam search, while DMMCS indicates the new proposed decoding method. For the latter, tags predicted by a medical image tagger are used.}
    \label{fig:per-modality-scores}
    \vspace{0.2cm}
\end{table*}

\section{Instructions}
\label{sec:instructions}

InstructBLIP \cite{instructblip} operates as a Vision-Language instruction-tuning model, which can adapt rapidly to new tasks based on specific instructions provided in its query prompt during training. Consequently, its performance depends largely on the quality of the instruction it is prompted with. While identifying the optimal instruction remains elusive, it is commonly agreed that concise and coherent instructions tend to yield better performance. However, due to the substantial memory demands of an InstructBLIP instance, conducting numerous experiments to find the optimal or near-optimal instruction can be very impractical. As a result, we conducted experiments using three different instruction prompts, which are presented in Table~\ref{table:instructios}.

\begin{table}[h]

    \centering
    \begin{tabular}{|p{0.5cm}|p{6.3cm}|}
     \hline
     \multicolumn{2}{|c|}{\textbf{InstructBLIP - Instruction Prompts}} \\
     \hline
     \centering \textbf{1} & {``Describe the given radiology image.''}\\
     \hline
     \centering \textbf{2} & {``You are an experienced radiologist. You are being given radiology images along with a brief medical diagnosis. Generate a descriptive caption that highlights the location, nature and severity of the abnormality of the radiology image.''}\\
     \hline
     \centering \textbf{3} & {``You are a helpful medical assistant. Generate a diagnostic report based on the patient's radiology examinations.''}\\
     \hline
    \end{tabular}
    \caption{\label{table:instructios} The three instructions that we defined in order to guide the InstructBLIP model throughout the DC task.}
\end{table}

\section{Per-modality scores}
 \label{per-modality}
 
 In this section, we conducted a modality-specific evaluation in order to assess and highlight any notable differences in performance between different modalities, which is presented in Table \ref{fig:per-modality-scores}. After our exploratory analysis, we observed that the ImageCLEFmedical 2023 dataset contains examinations originating from four main medical modalities, which are described in 
 % namely X-Ray, Computer Tomography, MRI and Ultrasonography.
 Thus, we split it into four subsets and evaluated the performance of the standard beam search and our proposed decoding method individually. In the rare case that an image belongs to more than one modality, we randomly select one of them to assign it. We observed that DMMCS outperforms beam search across almost all methods and modalities, remaining consistent with our findings on the entire dataset.

 \section{Perplexity}
\label{sec:perplexity-results}

We also provide perplexity scores (Table~\ref{tab:perplexity}) obtained using ClinicalT5 to measure the fluency of each model.\footnote{ We were granted access to ClinicalT5 through PhysioNet: \url{https://www.physionet.org/content/clinical-t5/1.0.0/},  {{L}ast accessed: 2024-06-05}.} ClinicalT5 is a biomedical version of T5 \cite{DBLP:journals/corr/abs-1910-10683}, pre-trained on the MIMIC-III dataset \cite{mimic-iii}. 
We observe that our proposed method generates more fluent captions than the baseline decoding methods, as it achieves lower perplexity scores in most cases.

\begin{table}[h!]
    \centering
    \begin{tabular}{|p{1.65cm}||p{0.62cm}|p{1.14cm}|p{1.14cm}|p{1cm}|}
        \hline
         \multicolumn{5}{|c|}{\small\textbf{ImageCLEFmedical 2023 Dataset - Predicted Tags}} \\
        \hline
        \multicolumn{1}{|c||}{\multirow{4}{*}{}} & \multicolumn{4}{c|}{\small\textbf{Perplexity}} \\
        \cline{2-5}
        & \centering\footnotesize \textit{BS} & \centering\footnotesize \textit{ConBS $\forall$} & \centering\footnotesize \textit{ConBS $\exists$} & \footnotesize \textit{DMMCS} \\
        \hline
        \footnotesize \centering \textbf{Show \& Tell ($\times10^4$)} & \footnotesize \centering  18.38 \hspace{0.15cm} & \footnotesize \centering 31.12 & \footnotesize \centering 18.75 & \hspace{0.15cm}\footnotesize \textbf{18.25} \\
        \hline
        \footnotesize \centering \textbf{ViT-GPT2 ($\times10^4$)} & \footnotesize \centering 17.70 & \footnotesize \centering 26.96 & \footnotesize \centering 20.30 & \hspace{0.15cm}\footnotesize \textbf{17.51} \\
        \hline
        \footnotesize \centering \textbf{InstructBLIP ($\times10^4$)} & \footnotesize \centering 16.47 & \footnotesize \centering 16.61 & \footnotesize \centering 22.12 &  \hspace{0.15cm}\footnotesize \textbf{15.95} \\
        \hline
        \footnotesize \centering \textbf{Flamingo ($\times10^4$)} & \footnotesize \centering 20.79 & \footnotesize \centering 23.69 & \footnotesize \centering 21.02 & \hspace{0.15cm}\footnotesize \textbf{20.62} \\
        % \footnotesize \centering \textbf{Show \& Tell} & \footnotesize \centering  3.053433e+06 \hspace{0.15cm} & \footnotesize \centering 3.030442e+06 & \footnotesize \centering \textbf{2.885060e+06} & \hspace{0.15cm}\footnotesize 3.010899e+06 \\
        % \hline
        % \footnotesize \centering \textbf{ViT-GPT2} & \footnotesize \centering 3.0112514e+07 & \footnotesize \centering 3.4069076e+07 & \footnotesize \centering 3.8352264e+07 & \hspace{0.15cm}\footnotesize \textbf{2.9885472e+07} \\
        % \hline
        % \footnotesize \textbf{InstructBLIP} & \footnotesize \centering 1.3987e+08 & \footnotesize \centering 1.2620e+08 & \footnotesize \centering 1.3551e+08 &  \hspace{0.15cm}\footnotesize \textbf{1.2500e+08} \\
        % \hline
        % \footnotesize \centering \textbf{Flamingo} & \footnotesize \centering \textbf{2.8383642e+07} & \footnotesize \centering 3.3160176e+07 & \footnotesize \centering 3.8551092e+07 & \hspace{0.15cm}\footnotesize 2.9019578e+07 \\
        \hline
    \end{tabular}
    \begin{tabular}{|p{1.65cm}||p{0.62cm}|p{1.14cm}|p{1.14cm}|p{1cm}|}
        \hline
         \multicolumn{5}{|c|}{\small\textbf{MIMIC-CXR Dataset - Predicted Tags}} \\
        \hline
        \multicolumn{1}{|c||}{\multirow{4}{*}{}} & \multicolumn{4}{c|}{\small\textbf{Perplexity}} \\
        \cline{2-5}
        & \centering\footnotesize \textit{BS} & \centering\footnotesize \textit{ConBS $\forall$} & \centering\footnotesize \textit{ConBS $\exists$} & \footnotesize \textit{DMMCS} \\
        \hline
        \footnotesize \centering \textbf{Show \& Tell ($\times10^6$)} & \footnotesize \centering  3.05 \hspace{0.15cm} & \footnotesize \centering 3.03 & \footnotesize \centering \textbf{2.88} & \hspace{0.15cm}\footnotesize 3.01 \\
        \hline
        \footnotesize \centering \textbf{ViT-GPT2 ($\times10^7$)} & \footnotesize \centering 3.01 & \footnotesize \centering 3.40 & \footnotesize \centering 3.83 & \hspace{0.15cm}\footnotesize \textbf{2.98} \\
        \hline
        \footnotesize \centering \textbf{InstructBLIP ($\times10^8$)} & \footnotesize \centering 1.39 & \footnotesize \centering 1.26 & \footnotesize \centering 1.35 &  \hspace{0.15cm}\footnotesize \textbf{1.25} \\
        \hline
        \footnotesize \centering \textbf{Flamingo ($\times10^7$)} & \footnotesize \centering \textbf{2.83} & \footnotesize \centering 3.31 & \footnotesize \centering 3.85 & \hspace{0.15cm}\footnotesize 2.90 \\
        % \footnotesize \centering \textbf{Show \& Tell} & \footnotesize \centering  3.053433e+06 \hspace{0.15cm} & \footnotesize \centering 3.030442e+06 & \footnotesize \centering \textbf{2.885060e+06} & \hspace{0.15cm}\footnotesize 3.010899e+06 \\
        % \hline
        % \footnotesize \centering \textbf{ViT-GPT2} & \footnotesize \centering 3.0112514e+07 & \footnotesize \centering 3.4069076e+07 & \footnotesize \centering 3.8352264e+07 & \hspace{0.15cm}\footnotesize \textbf{2.9885472e+07} \\
        % \hline
        % \footnotesize \textbf{InstructBLIP} & \footnotesize \centering 1.3987e+08 & \footnotesize \centering 1.2620e+08 & \footnotesize \centering 1.3551e+08 &  \hspace{0.15cm}\footnotesize \textbf{1.2500e+08} \\
        % \hline
        % \footnotesize \centering \textbf{Flamingo} & \footnotesize \centering \textbf{2.8383642e+07} & \footnotesize \centering 3.3160176e+07 & \footnotesize \centering 3.8551092e+07 & \hspace{0.15cm}\footnotesize 2.9019578e+07 \\
        \hline
    \end{tabular}
    \begin{tabular}{|p{1.65cm}||p{0.62cm}|p{1.14cm}|p{1.14cm}|p{1cm}|}
    \hline
    \hline
    \centering \footnotesize \textit{\textbf{Wins}} & \centering\footnotesize 1 & \centering\footnotesize 0 & \centering\footnotesize 1  & \hspace{0.47cm}\footnotesize \textbf{6} \\
        \hline
    \end{tabular}
    \caption{The perplexity of each model and decoding method on the ImageCLEFmedical 2023 and the MIMIC-CXR dataset computed using ClinicalT5.}
    \label{tab:perplexity}
\end{table}

\section{Clinical Accuracy - Extensive Results}
\label{sec:ca-results}

Table \ref{fig:ca} presents an extensive evaluation of the Clinical Accuracy measure across all models and decoding methods. We share more details on how we calculate the CA between the gold and generated captions in Section ~\ref{sec:4.4}.

 \begin{table}[h!]
    \centering
    \begin{tabular}{|p{1.65cm}||p{0.62cm}|p{1.14cm}|p{1.14cm}|p{1cm}|}
        \hline
         \multicolumn{5}{|c|}{\small\textbf{ImageCLEFmedical 2023 - With Predicted Tags}} \\
        \hline
        \multicolumn{1}{|c||}{\multirow{4}{*}{}} & \multicolumn{4}{c|}{\small\textbf{Clinical Accuracy}} \\
        \cline{2-5}
        & \centering\footnotesize \textit{BS} & \centering\footnotesize \textit{ConBS $\forall$} & \centering\footnotesize \textit{ConBS $\exists$} & \footnotesize \textit{DMMCS} \\
        \hline
        \footnotesize \centering \textbf{Show \& Tell} & \footnotesize \centering \textbf{92.03} & \footnotesize \centering 90.82 & \footnotesize \centering 91.73 &  \hspace{0.15cm}\footnotesize {91.38}\\
        \hline
        \footnotesize \centering \textbf{ViT-GPT2} & \footnotesize \centering 91.55 & \footnotesize \centering 90.51 & \footnotesize \centering \textbf{91.77} & \hspace{0.15cm}\footnotesize 91.60 \\
        \hline
        \footnotesize \textbf{InstructBLIP} & \footnotesize \centering 91.47 & \footnotesize \centering 88.00 & \footnotesize \centering 90.86 &  \hspace{0.15cm}\footnotesize \textbf{91.73} \\
        \hline
        \footnotesize \centering \textbf{Flamingo} & \footnotesize \centering \textbf{91.29} & \footnotesize \centering 90.86 & \footnotesize \centering 89.82 & \hspace{0.15cm}\footnotesize 90.99 \\
        \hline
    \end{tabular}
    \\
    \begin{tabular}{|p{1.65cm}||p{0.62cm}|p{1.14cm}|p{1.14cm}|p{1cm}|}
        \hline
         \multicolumn{5}{|c|}{\small\textbf{MIMIC-CXR Dataset - With Predicted Tags}} \\
        \hline
        \multicolumn{1}{|c||}{\multirow{4}{*}{}} & \multicolumn{4}{c|}{\small\textbf{Clinical Accuracy}} \\
        \cline{2-5}
        & \centering\footnotesize \textit{BS} & \centering\footnotesize \textit{ConBS $\forall$} & \centering\footnotesize \textit{ConBS $\exists$} & \footnotesize \textit{DMMCS} \\
        \hline
        \footnotesize \centering \textbf{Show \& Tell} & \footnotesize \centering 83.87 & \footnotesize \centering \textbf{92.73} & \footnotesize \centering 84.51 &  \hspace{0.15cm}\footnotesize 84.80\\
        \hline
        \footnotesize \centering \textbf{ViT-GPT2} & \footnotesize \centering 84.77 & \footnotesize \centering \textbf{91.78} & \footnotesize \centering 87.91 & \hspace{0.15cm}\footnotesize 84.74 \\
        \hline
        \footnotesize \textbf{InstructBLIP} & \footnotesize \centering 83.33 & \footnotesize \centering 80.03 & \footnotesize \centering 77.45 &  \hspace{0.15cm}\footnotesize \textbf{84.19} \\
        \hline
        \footnotesize \centering \textbf{Flamingo} & \footnotesize \centering 80.12 & \footnotesize \centering \textbf{88.82} & \footnotesize \centering 85.82 & \hspace{0.15cm}\footnotesize 79.85 \\
        \hline
    \end{tabular}
    \begin{tabular}{|p{1.65cm}||p{0.62cm}|p{1.14cm}|p{1.14cm}|p{1cm}|}
    \hline
    \hline
    \centering \footnotesize \textit{\textbf{Wins}} & \centering\footnotesize 2 & \centering\footnotesize \textbf{3} & \centering\footnotesize 1  & \hspace{0.47cm}\footnotesize 2 \\
        \hline
    \end{tabular}
    \caption{The clinical accuracy of each model and decoding method on the ImageCLEFmedical 2023 and MIMIC-CXR datasets.}
    \label{fig:ca}
\end{table}

\end{document}